%% file: main.tex
\documentclass[11pt, a4paper, copyright, gdm]{google}

\input{macros_imports}

\usepackage[authoryear, sort&compress, round]{natbib}
\title{S1-NexusAgent:\\a Self-Evolving Agent Framework for Multidisciplinary Scientific Research}

\renewcommand{\today}{}
\author{S1-NexusAgent Team, Institute of Automation, Chinese Academy of Sciences}
\let\cite\citep

\begin{abstract}
Modern scientific research relies on large-scale data, complex workflows, and specialized tools, which existing LLMs and tool-based agents struggle to handle due to limitations in long-horizon planning, robust goal maintenance, and continual learning from execution. To address these issues, in this work, we propose S1-NexusAgent, a self-evolving agent framework designed for multidisciplinary scientific research. S1-NexusAgent adopts a hierarchical Plan-and-CodeAct execution paradigm, decoupling global scientific planning from subtask-level tool execution through a dual-loop architecture, thereby enabling stable modeling of complex research workflows. The system natively supports the Model Context Protocol (MCP), integrates up to thousands of cross-disciplinary scientific tools, and achieves efficient orchestration of heterogeneous research tools via intention-aware dynamic tool retrieval and hot-plug mechanisms. To address long-context and large-scale data challenges in scientific settings, S1-NexusAgent introduces object-reference-based sparse context management, which enables sub-task context isolation and intermediate result compression. Building on this, a Critic Agent automatically evaluates complete execution trajectories and distills high-quality research paths into reusable Scientific Skills, forming a closed loop for continuous self-evolution, which is valuable for sustainable and long-horizon scientific research. Experiments on authoritative scientific benchmarks involving long-horizon planning and complex specialized tool orchestration, including biomini-eval (biology), ChemBench (chemistry), and MatSciBench (material science), demonstrate that S1-NexusAgent achieves state-of-the-art performance, validating its effectiveness and generalization capability in complex scientific tasks. The code of S1-NexusAgent is available at \url{https://github.com/CASIA-LM/S1-NexusAgent}.

\end{abstract}

\begin{document}

\maketitle

\subimport{1_introduction}{1_introduction}

\input{2_related_work}

\section{Architecture}
\label{sec:architecture}
\subimport{3_architecture_sections}{1_overview}
\subimport{3_architecture_sections}{2_intent_and_tool_selection}
\subimport{3_architecture_sections}{3_dual_loop}
% \subimport{2_architecture_sections}{4_CodeAct_and_MCP}
\subimport{3_architecture_sections}{4_context_management}

\subimport{3_architecture_sections}{5_self-evolution}

\section{Agentic Training}
\label{sec:training}
\subimport{4_agentic_training}{1_SFT}

\subimport{4_agentic_training}{2_RL}

\section{Quantitative evaluation}
\label{sec:results_quantitative}
\subimport{5_results_sections}{1_setups}
\subimport{5_results_sections}{3_Biomni_eval}
\subimport{5_results_sections}{2_ChemBench}
\subimport{5_results_sections}{4_MatSciBench}

\section{Case studies of S1-NexusAgent}
\label{sec:example_use_cases}
\subimport{6_use_cases_sections}{1_use_cases}

\input{7_discussion}

\input{8_conclusion}

\clearpage

% \iffalse
\section*{Team Members}
\label{sec:team_member}
\begin{itemize}

\item \textbf{First Authors}:

Ziliang Pang (CASIA)

*Shuo Ren (CASIA)

\item \textbf{Core Contributors}:

Zhang Zhou (CASIA)

Xiaomian Kang (CASIA)

Qianlong Du (CASIA)

Can Xie (CASIA)

Ziyi Zhang (CASIA)

Zhenjiang Ren (CASIA)

\item \textbf{Contributors (listed in alphabetical order by last name)}:

Yingying Chen (CASIA)

Guangchuan Guo (Taichu)

Longteng Guo (CASIA)

Wenjing Li (CASIA)

Zhiyuan Li (CASIA)

Dan Liu (CASIA)

Xiaotao Wang (CASIA)

Yuying Wang (Taichu)

Chunpeng Yang (Taichu)

Ge Yang (CASIA)

Haisheng Yu (Taichu)

Zhidong Zhang (Taichu)

Yang Zhao (CASIA)

Bingke Zhu (CASIA)

\item \textbf{Supervisors (listed in alphabetical order by last name)}:

Jian Cheng (CASIA)

Rufeng Chu (Taichu)

Linjing Li (CASIA)

Jing Liu (CASIA)

Jinqiao Wang (CASIA)

Ge Yang (CASIA)

Zhaoxiang Zhang (CASIA)

Dajun Zeng(CASIA)

Chengqing Zong (CASIA)

\item \textbf{Project Lead}:

*Jiajun Zhang (CASIA)
\vspace{1em}

(*corresponding authors)

\end{itemize}
% \fi

CASIA: Institute of Automation, Chinese Academy of Sciences.

Taichu: Zhongke Zidong Taichu (Beijing) Technology Co., Ltd.

\clearpage
\bibliography{main}
\clearpage

\section{Appendix}
\label{sec:appendix}
\subimport{9_appendix}{biomni_example}
\subimport{9_appendix}{ChemBench_example}

\subimport{9_appendix}{matscibench_example}

\end{document}

%% file: macros_imports.tex
\usepackage{float}
\usepackage{makecell}        % For advanced multi-line cells (\thead, \makecell)
\usepackage{fontawesome5}  % For icons
\usepackage{xspace}
\usepackage{multicol}
\usepackage{multirow}
\usepackage{longtable}
\usepackage{subcaption}
\usepackage{import}
\usepackage[usestackEOL]{stackengine}
\usepackage{framed}

\definecolor{headerbg}{RGB}{25, 25, 25}       % Black header background
\definecolor{gridgray}{gray}{0.85}           % Light gray for table lines
\definecolor{goodgreen}{RGB}{28, 142, 64}    % Green for improvements
\definecolor{badred}{RGB}{237, 34, 39}       % Red for regressions
\definecolor{cclnotreachedcolor}{RGB}{220, 53, 69} 

\definecolor{TargetGreen}{RGB}{46, 139, 87}

\definecolor{TargetGreen}{RGB}{46, 139, 87}

\newcolumntype{C}{>{\centering\arraybackslash}X}

%% file: 1_introduction/1_introduction.tex
\section{Introduction}
\label{sec:introduction}
Scientific research is undergoing a profound paradigm shift. From high-throughput omics analysis to large-scale materials screening, and from astronomical survey data processing to complex physical simulations, research workflows are rapidly transitioning from traditional experience-driven practices toward highly data-driven and computation-intensive paradigms. However, this transformation has also substantially increased the complexity of scientific tasks: research problems frequently span multiple disciplines, experimental trajectories critically depend on intermediate feedback, and effective progress requires the coordination of numerous specialized tools and heterogeneous data sources.

In recent years, large language models (LLMs) and their derived intelligent agents have shown initial promise in scientific reasoning and tool invocation in special domains~\cite{kang2023multi, boiko2023autonomous, M.Bran2024, hu2025electromagnetic, xia2025naturelm, zhang2025exploring, szymanski2023autonomous, gottweis2025aicoscientist, huang2025biomni,  zhang2025origene, jin2025stella, jaiswal2024improving, moss2025ai, wang2025starwhisper, schmidgall2025agent, zhang2025intern, wen2026innovator, bai2025intern, ren2026sci}. Nevertheless, most existing approaches are tailored to specific domains or workflows, which limits their generality and flexibility. Some are designed to support multiple disciplines, but still face challenges in real-world scientific settings. First, long-horizon planning capabilities are limited: linear or myopic planning mechanisms struggle to support multi-stage, long-term scientific exploration and often drift from research objectives. Second, tool operation lacks flexibility: most tool-calling frameworks rely on static interfaces or fixed prompt templates, making them poorly suited to the rapidly evolving scientific tool ecosystem. Third, context management poses a significant challenge: the large-scale data and long temporal reasoning inherent in scientific tasks frequently cause context expansion, severely constraining agent stability and scalability. Fourth, continual evolution remains insufficient: scientific research is inherently iterative, involving trial-and-error, repeated experimentation, and refinement, which demands strong self-evolving capabilities for agents to sustain long-term research efficiency and robustness. 

We argue that an intelligent scientific agent not only gives answers, but operates like a scientist, i.e., planning experiments, executing analyses, reflecting on results, and continuously learning from experience. Motivated by this principle, we propose \textbf{S1-NexusAgent}, a general-purpose scientific agent framework. \emph{\textbf{Nexus}} emphasizes the framework’s role as a unifying hub that connects and supports scientific workflows across multiple disciplines. The framework is built with a \textbf{Plan-and-CodeAct} paradigm, which employs hierarchical planning to decompose high-level research objectives into executable sub-tasks, and leverages CodeAct \cite{wang2024executable} to tightly couple reasoning with code execution, enabling the agent to manipulate real scientific tools programmatically. Together, these design naturally gives rise to an inner–outer dual-loop architecture, which explicitly models the iterative experimentation and strategy refinement inherent in scientific exploration.

% TODO: Tool Integration \& MCP
At the execution level, \textbf{S1-NexusAgent} adopts \textbf{CodeAct} as a unified action paradigm, enabling the agent to interact with real scientific tools through executable code rather than static textual descriptions. This design explicitly couples scientific reasoning, tool selection, and execution into a single, inspectable process. The system defines a standardized tool interface specification tailored for scientific scenarios, which provides a unified abstraction over tool functional semantics, input--output specifications, and execution constraints. Based on this specification, S1-NexusAgent has integrated and managed up to thousands of scientific tools spanning multiple domains, including but not limited to life sciences, chemistry, materials science, astronomy, mathematics, and scientific computing. In addition to the native tool interface, the framework supports the MCP as an extensible layer for external tool access. To mitigate reasoning interference caused by large tool collections, S1-NexusAgent introduces an intent-aware dynamic tool retrieval and on-demand injection mechanism, which selectively loads only a small set of tools that are highly relevant to the current sub-task into the execution context. This design enables the framework to support a large-scale, multi-domain tool ecosystem while preserving reasoning accuracy and system robustness.

% TODO: Sparse Context Management
Scientific tasks commonly involve large-scale data dependencies, long-horizon reasoning, and iterative exploration across multiple sub-tasks. To address these challenges, S1-NexusAgent constructs a \textbf{sparse context management framework} composed of four complementary mechanisms, including \emph{object-level context referencing}, \emph{sub-task-level context isolation}, \emph{execution trajectory compression}, and \emph{planning-aware context augmentation}. This framework decouples raw large-scale scientific data from high-value decision-relevant information, enabling key scientific evidence to remain traceable and reusable while effectively suppressing context expansion and reasoning noise. As a result, S1-NexusAgent maintains reasoning coherence, planning stability, and system scalability throughout cross-stage and long-duration scientific exploration processes.

Furthermore, S1-NexusAgent is not a static system. By introducing a \textbf{T}rajectory-\textbf{E}valuation-based \textbf{S}elf-\textbf{E}volution mechanism (\textbf{TE-SE}), the framework automatically identifies high-quality scientific execution paths and distills them into reusable Scientific Skills. Through the continual acquisition and reuse of such skills across tasks, the agent accumulates research experience and progressively improves its decision-making efficiency over time.

We evaluate our framework on three widely used scientific benchmarks that require long-horizon planning and complex orchestration of specialized tools, including biomni-eval (biology), ChemBench (chemistry), and MatSciBench (materials science). The results show that S1-NexusAgent achieves state-of-the-art performance across these benchmarks, demonstrating its effectiveness and generalization capability in complex, real-world scientific tasks. Beyond benchmark evaluation, S1-NexusAgent has also been applied in several real-world research settings, where it supports end-to-end scientific workflows involving iterative analysis, tool coordination, and long-term task execution, further indicating its practical utility in realistic scientific scenarios.

The main contributions of this work are summarized as follows:

\begin{itemize}
    \item We propose a general-purpose scientific agent architecture based on the Plan-and-CodeAct paradigm, which stably supports long-horizon scientific tasks through an inner–outer dual-loop mechanism.
    \item We build a dynamic scientific tool ecosystem with native support for the Model Context Protocol (MCP), enabling efficient retrieval and hot-plugging of large-scale heterogeneous scientific tools.
    \item We design a sparse context management mechanism tailored for scientific tasks, effectively addressing the challenges of long textual contexts and large-scale data.
    \item We introduce a trajectory-evaluation-based self-evolution framework that distills successful scientific execution paths into reusable Scientific Skills, enabling continuous growth of agent capabilities.
\end{itemize}

%% file: 2_related_work.tex
\section{Related Work}
\label{sec:related_work}
\subsection{Domain-Specific Scientific Agents}
A large body of recent work builds domain-specialized scientific agents that encode substantial task priors for a particular field, such as molecular design and optimization, materials discovery/characterization, and biomedical discovery workflows \cite{ren2026sci}. Representative systems include chemistry and materials agents like ChemCrow \cite{M.Bran2024}, CACTUS \cite{mcnaughton2024cactus}, IR-Agent \cite{noh2025ir}, HoneyComb \cite{zhang-etal-2024-honeycomb}, MatPilot \cite{ni2024matpilot}, MAPPS \cite{zhou2025toward}, and others, as well as life-science agents spanning drug discovery, genomics/bioinformatics, and protein/CRISPR applications (e.g., DrugAgent \cite{inoue2024drugagent}/DrugPilot \cite{li2025drugpilot}, Biomni \cite{huang2025biomni}, BioDiscoveryAgent \cite{roohani2024biodiscoveryagent}, CellVoyager \cite{alber2025cellvoyager}, GeneGPT \cite{jin2024genegpt}, CRISPR-GPT \cite{huang2024crispr}, etc.). These agents often achieve strong performance by aligning planning and tool use with a narrowly defined workflow or domain toolchain. 

While domain-specific agents can be highly effective within a fixed scope, S1-NexusAgent targets cross-domain scientific tasks with a unified architecture, emphasizing long-horizon stability, flexible tool orchestration over a large heterogeneous tool ecosystem, and system-level mechanisms (e.g., dual-loop execution, context structuring, and self-evolution) that are not tied to a single domain’s workflow.

\subsection{Complex General Scientific Agent Systems}
Beyond domain-specific scientific agents, a small line of work targets multidisciplinary scientific agent systems that aim to support scientific problem solving across multiple disciplines within a unified framework. Unlike agents designed for a single field or workflow, these systems emphasize domain-agnostic architectural mechanisms, relying on dynamic tool use rather than fixed pipelines to adapt to different scientific contexts. Representative examples include InternAgent \cite{zhang2025intern}, which coordinates heterogeneous agents and tools to handle cross-domain scientific tasks, and SciMaster \cite{chai2025scimaster}, which explores general-purpose scientific intelligence through scalable, tool-augmented agentic workflows evaluated across diverse scientific domains, such as materials science \cite{zhang2025bohrium}, physics \cite{miao2025physmaster}, machine learning \cite{zhu2026toward}, etc. 

While InternAgent and SciMaster highlight the feasibility of multidisciplinary scientific agents, they largely rely on monolithic or loosely structured control loops. In contrast, S1-NexusAgent explicitly introduces an \textbf{inner–outer dual-loop} architecture that separates long-horizon, task-level planning from subtask-level executable reasoning. This separation enables S1-NexusAgent to maintain global goal stability while supporting localized trial-and-error, systematic replanning, and trajectory-level evaluation, providing a clearer structural basis for long-term robustness and self-evolution through reusable Scientific Skills.

\subsection{Code-enhanced and programmatic scientific agents}
Another line of related work emphasizes code as a first-class action and planning medium for scientific agents. Rather than relying solely on natural-language reasoning, these systems generate executable artifacts—such as programs, scripts, or structured workflows—to drive scientific computation, data analysis, and experimental validation, improving precision, reproducibility, and controllability. Representative systems include AlphaEvolve \cite{novikov2025alphaevolvecodingagentscientific}, which performs iterative program synthesis and evaluation for scientific optimization; workflow-oriented agents such as AIGS \cite{liu2024aigs} and ORGANA \cite{darvish2025organa}, which synthesize executable pipelines for scientific processes; and tool-augmented analysis agents such as Chemist-X \cite{chen2023chemist} and Biomni \cite{huang2025biomni}, which iteratively generate and execute code to validate hypotheses and refine results. In these systems, planning is often encoded implicitly through program structure or explicitly via intermediate program representations, with execution serving as the primary grounding mechanism. 

S1-NexusAgent aligns with the code-centric philosophy of these approaches but differs in key aspects. It adopts CodeAct \cite{wang2024executable} as a unified execution interface, explicitly coupling reasoning and code execution within an iterative inner loop rather than treating code generation as a one-off artifact. Moreover, code execution is embedded within an explicit inner–outer dual-loop architecture, where programmatic actions support local hypothesis validation while global planning and strategy revision occur at the task level. Finally, S1-NexusAgent incorporates trajectory evaluation and Scientific Skill distillation, enabling successful code-driven executions to be accumulated and reused—extending code-enhanced agents from execute-to-solve toward execute-to-learn-and-reuse in long-horizon scientific workflows.

%% file: 3_architecture_sections/1_overview.tex
\subsection{Overview}
\label{sec:overview}

\begin{figure*}[!htp]
    \centering
    \includegraphics[width=\textwidth]{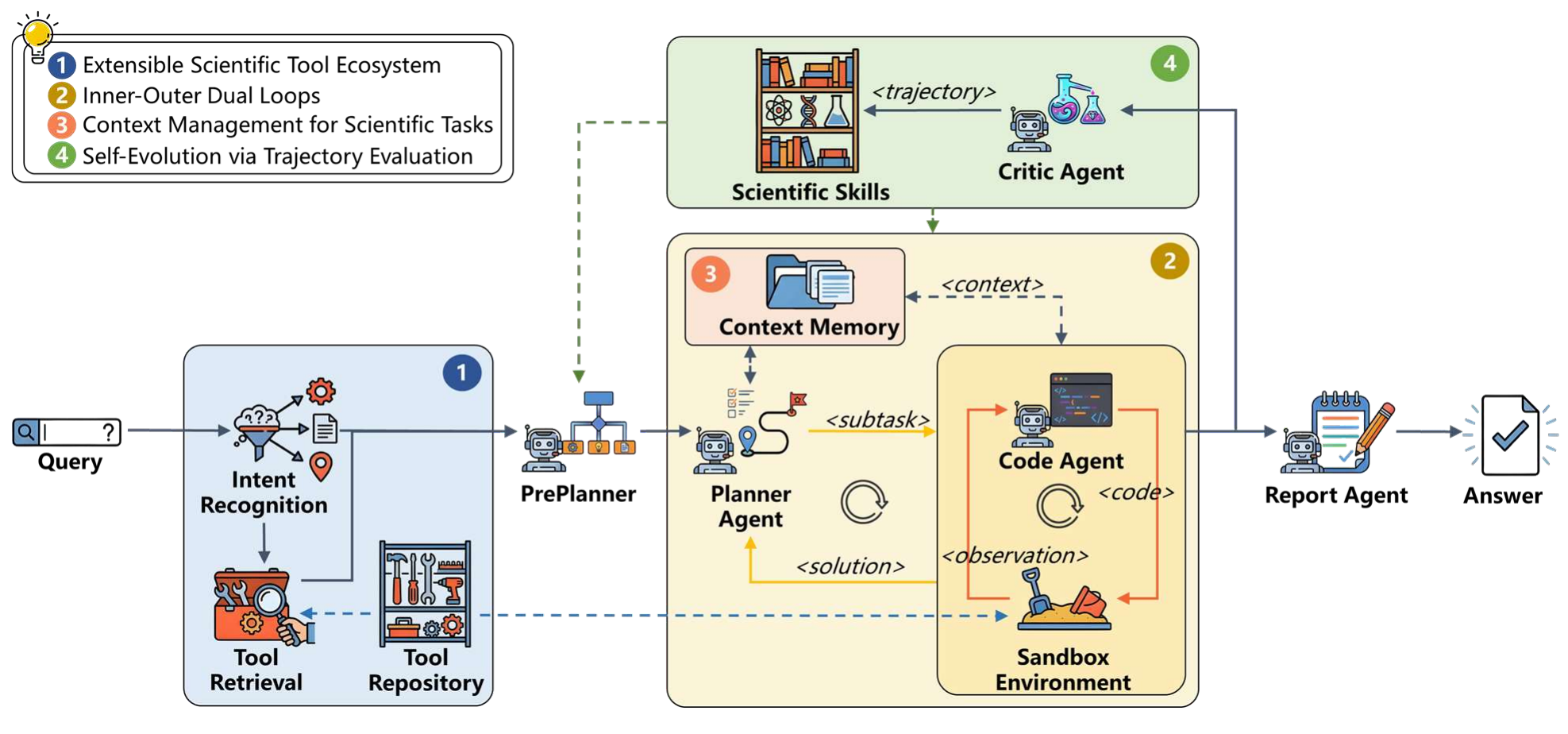}
    \caption{Architectural Overview of S1-NexusAgent.}
    \label{fig:architecture}
\end{figure*}

S1-NexusAgent is designed for real-world scientific tasks that are highly complex, long-horizon, and characterized by uncertain and iterative feedback. It adopts a general-purpose scientific agent architecture centered on the \textbf{Plan-and-CodeAct} paradigm (as illustrated in Figure \ref{fig:architecture}). Through multi-agent collaboration and an \textbf{inner–outer dual-loop mechanism}, the system progressively transforms high-level scientific intents into executable experimental and analytical workflows, forming a closed loop of understanding, planning, execution, reflection, and evolution.

When a user submits a high-level scientific query (e.g., “analyze the potential mechanisms by which a gene mutation leads to drug resistance”), the system first invokes an \textbf{Intent Recognition Agent} to parse the request and map the natural language input into structured scientific objectives. These objectives capture key elements such as the research subject, task type, relevant domains, and tool-selection cues that guide subsequent execution. The structured intent is then passed to the \textbf{Tools Retrieval} module, which constrains the  tool space available for downstream planning and execution.

Once the tool space is determined, a \textbf{Pre-Planner Agent} generates a global task outline based on the current research objective. Rather than specifying concrete execution details, this outline characterizes high-level research stages and sub-goals, analogous to the research roadmap formed by human scientists before conducting experiments. This pre-planning mechanism effectively mitigates the research goal drift commonly observed in long-horizon tasks.

The system then enters a dynamic planning phase governed by the Planner Agent. The Planner continuously maintains the current task state and determines at each step whether further sub-task decomposition is required or whether the task can be considered complete. When concrete execution is needed, the Planner dispatches the corresponding sub-task to the CodeAct module; when the research objective is deemed satisfied, it terminates execution and proceeds to result aggregation. This decision-making process constitutes the \textbf{outer loop} of S1-NexusAgent, which is responsible for global control over the scientific workflow.

Complementing this task-level planning, S1-NexusAgent introduces an \textbf{inner loop} centered on CodeAct during sub-task execution. Once a sub-task is assigned to CodeAct, a \textbf{Code Agent} orchestrates and invokes scientific tools within a sandbox environment using a tightly coupled reasoning-and-coding paradigm, iteratively refining its actions based on returned observations. This inner loop continues until the sub-task achieves its intended  objective or is judged as no longer improvable. The final execution outcome is then fed back to the Planner Agent, which updates the global task state and informs subsequent decisions.

The coordination between the inner and outer loops enables S1-NexusAgent to maintain global goal coherence while allowing flexible local exploration and iterative refinement at the sub-task level. This dual-loop design supports trial-and-error experimentation without sacrificing long-term task consistency, making the system well-suited for complex scientific problem that requires both strategic planning and adaptive execution.

In the following sections, we will introduce each core module in detail. 

%% file: 3_architecture_sections/2_intent_and_tool_selection.tex
\subsection{An Extensible Scientific Tool Ecosystem}
\label{sec:tool_retrieval}
% To cope with a large and continuously growing tool repository, S1-NexusAgent introduces a mechanism of intention-aware \textbf{Dynamic Hot-Plugging (DHP)} for tool retrieval. The Tools Retrieval module first performs domain-level filtering based on the structured intent produced by the Intent Recognition component, excluding tool categories that are irrelevant to the current scientific problem. Within the selected candidate domains, the system further conducts fine-grained semantic matching by incorporating the current sub-task context and dynamically injects a small subset of the most relevant tools into the following execution context.

% This on-demand injection and plug-and-play strategy substantially reduces contextual noise, prevents irrelevant tools from interfering with the reasoning process, and improves the accuracy and stability of tool invocation. As a result, S1-NexusAgent maintains high flexibility while achieving execution reliability suitable for real-world scientific research.

S1-NexusAgent addresses the challenge of managing vast scientific workflows by coupling an intelligent retrieval engine with a standardized, modular tool ecosystem. This structure allows the agent to navigate thousands of specialized instruments without overwhelming its reasoning context.

\subsubsection{Intent-Aware Dynamic Hot-Plugging (DHP)}
To navigate the large-scale tool repository efficiently, S1-NexusAgent utilizes an Intent-Aware Dynamic Hot-Plugging (DHP) mechanism. This process ensures that only the tools most relevant to the immediate scientific problem are loaded into the execution context, preventing "context explosion" and reducing reasoning noise.

The retrieval pipeline operates in three stages as shown in Figure \ref{fig:tool_retrieval}:
% --- Figure 1: Tool Retrieval ---
\begin{figure*}[!htp]
    \centering
    \includegraphics[width=0.85\textwidth]{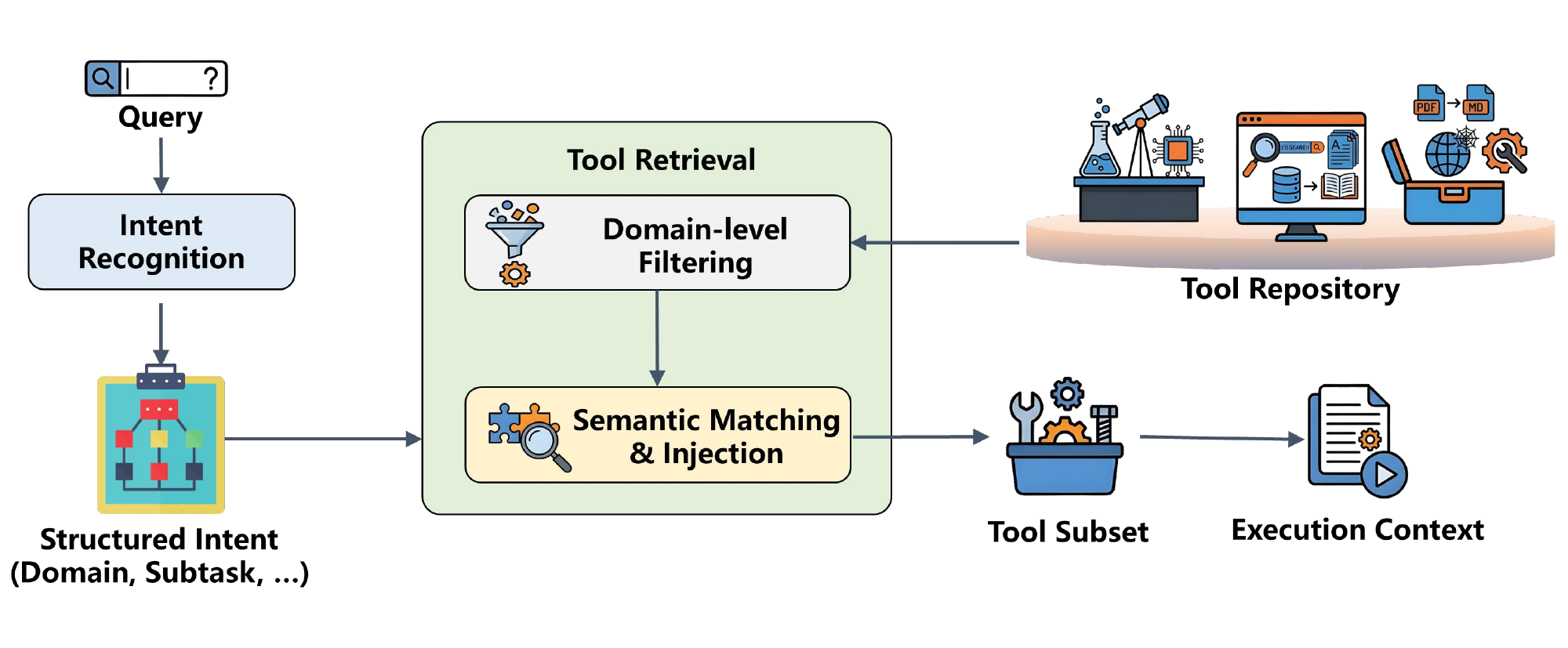}
    \caption{\textbf{Overview of the Intent-aware Dynamic Hot-Plugging (DHP) mechanism for scalable tool retrieval.} The framework first performs domain-level filtering based on structured intent to prune irrelevant tool categories. Subsequently, fine-grained semantic matching is conducted within candidate domains to facilitate the plug-and-play injection of the most relevant tools into the execution context. This on-demand strategy effectively mitigates contextual noise and prevents reasoning interference, ensuring robust tool invocation in large-scale scientific toolsets.}
    \label{fig:tool_retrieval}
\end{figure*}

\paragraph{(1) Intent Recognition.}When a user provides a query, an Intent Recognition Agent parses the natural language request into structured scientific objectives, identifying key variables such as the research domain, task type, and specific sub-goals.

\paragraph{(2) Domain-Level Filtering.} Using this structured intent, the system first filters out tool categories that are fundamentally irrelevant to the current task (e.g., excluding astronomical data processing tools during a drug discovery task).

\paragraph{(3) Semantic Matching \& Injection.}Within the remaining candidate domains, the system performs fine-grained semantic matching against the current sub-task context. It then dynamically "hot-plugs" this curated subset of tools into the active environment, making them immediately available for the agent's use.

\subsubsection{The Scientific Tool Ecosystem and Execution Interface}

% --- Figure 5: Tool Ecosystem ---
\begin{figure*}[!htp]
    \centering
    \includegraphics[width=0.85\textwidth]{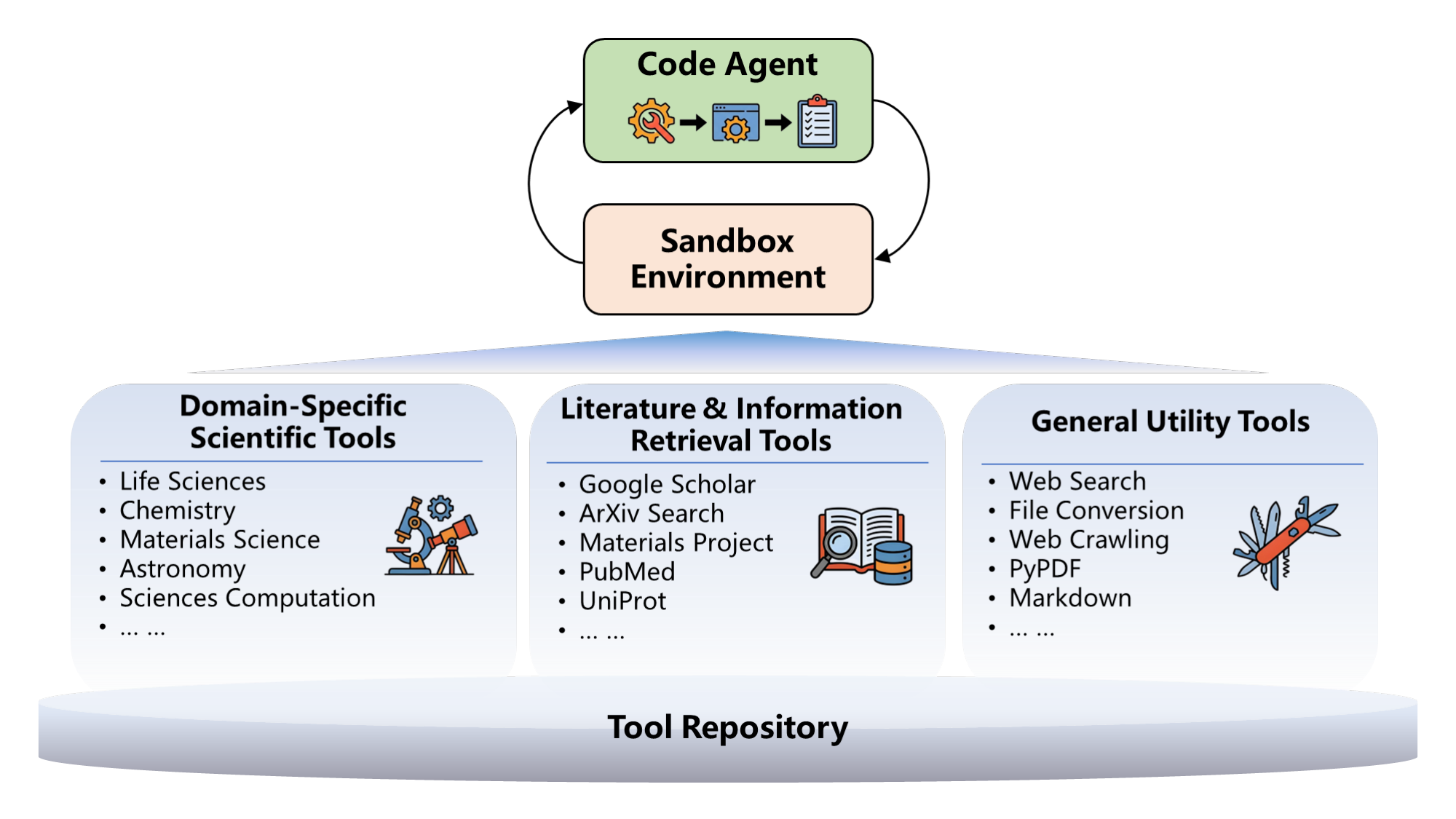}
    \caption{\textbf{The modular and extensible scientific tool ecosystem of S1-NexusAgent.} Scientific capabilities are abstracted into three functional layers: domain-specific tools (e.g., bioinformatics, materials simulation), literature retrieval services, and general utilities. By adopting CodeAct as the unified action paradigm and maintaining compatibility with standardized tool protocols, S1-NexusAgent provides a secure, sandboxed environment for programmatic tool interaction, enabling seamless integration of diverse scientific instruments and databases without model retraining.}
    \label{fig:tool_ecosystem}
\end{figure*}

Once the relevant tools are retrieved, S1-NexusAgent interacts with them through a unified interface designed for safety and reproducibility. The framework natively supports the Model Context Protocol (MCP), allowing it to integrate diverse resources into a standardized format. The ecosystem can currently manage up to thousands of tools categorized into three functional classes:

\paragraph{(1) Domain-Specific Scientific Tools.}Specialized instruments for fields like life sciences, chemistry, materials science, and astronomy, as well as mathematical packages for numerical computation.

\paragraph{(2) Literature and Information Retrieval Tools.}Services that connect the agent to external knowledge bases (e.g., Google Scholar, PubMed, UniProt), preventing isolated reasoning by providing access to up-to-date academic literature.

\paragraph{(3) General Utility Tools.}Auxiliary functions such as file conversion, web crawling, and data formatting that support the construction of end-to-end research pipelines.

To utilize these tools, the agent employs the CodeAct paradigm in the inner-loop (see Section \ref{sec:dualloop}). Rather than relying on rigid text-based API calls, the agent writes and executes code within a secure Sandbox Environment, coupling scientific reasoning with action, allowing the agent to perform complex data analysis,  tool parameter modification, and error correction programmatically.
% Through standardized tool interfaces, tool functionality descriptions, input–output schemas, and execution constraints can be automatically interpreted and invoked by the agent. This design enables continuous expansion of the tool ecosystem without requiring retraining of the core models, thereby supporting long-term scalability and extensibility of the system.

%% file: 3_architecture_sections/3_dual_loop.tex
\subsection{Unifying Global Planning and Local Exploration via Inner–Outer Dual Loops}
\label{sec:dualloop}
The core architectural advantage of S1-NexusAgent lies not only in the introduction of an \textbf{inner–outer dual-loop} structure, but more importantly in the synergy and complementarity between the two loops. The outer and inner loops are explicitly differentiated in terms of responsibilities, temporal scales, and optimization objectives, jointly forming a scientific decision-making system that maintains long-term goal coherence while supporting localized exploration and trial-and-error.

% --- Figure 2: Inner-Outer Dual Loops ---
\begin{figure*}[!htp]
    \centering
    \includegraphics[width=0.75\textwidth]{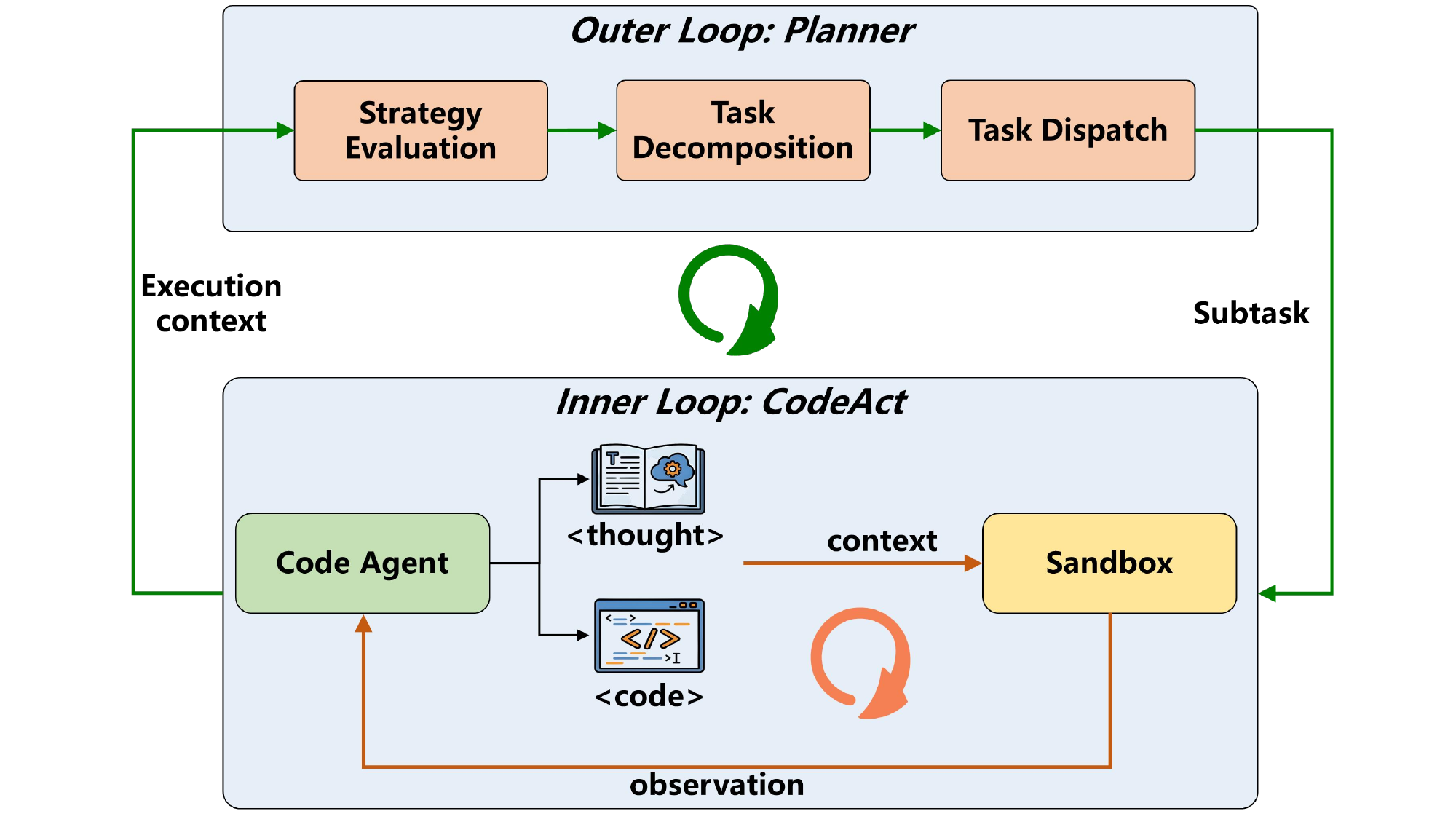}
    \caption{\textbf{The Inner--Outer Dual-Loop architecture of S1-NexusAgent.} The \textbf{Outer Loop}, governed by the Planner Agent, maintains global research objectives and regulates high-level strategy through recursive task decomposition. Complementing this, the \textbf{Inner Loop} leverages the CodeAct paradigm for localized exploration, enabling high-frequency trial-and-error, self-correction, and parameter tuning within specific subtasks. The synergy between these loops allows the system to adaptively scale execution depth and planning granularity according to task complexity.}
    \label{fig:dual_loop}
\end{figure*}

\textbf{The outer loop}, governed by the Planner Agent, operates at the task and stage levels and is responsible for maintaining global research objectives and regulating overall progress. Rather than focusing on low-level tool execution details, the Planner continuously assesses whether the current research should be further decomposed, redirected, or terminated based on structured feedback from completed sub-tasks. This design allows the system to perform strategy-level adjustments in response to uncertain experimental outcomes or intermediate failures, instead of rigidly following a predefined execution path.

Complementing this process, \textbf{the inner loop}, centered on CodeAct, operates at the sub-task and execution levels and focuses on validating specific scientific hypotheses and optimizing operational decisions. Within the inner loop, the Code Agent repeatedly invokes scientific tools through a tightly coupled reasoning + code paradigm, performing self-correction, parameter tuning, and tool substitution based on real-time observations until the subtask reaches a locally optimal feasible solution or is deemed infeasible. This process enables high-frequency trial-and-error and localized exploration without disrupting the global task structure.

Owing to the coordinated adjustment of task decomposition granularity and execution depth across the two loops, the architecture exhibits natural task-difficulty adaptivity. For well-structured scientific problems with simple dependencies, the Planner Agent often dispatches only a small number of (or even a single) sub-tasks, and CodeAct correspondingly completes execution with minimal internal iteration, enabling rapid convergence. In contrast, for complex problems with strong dependencies on intermediate feedback, the Planner progressively issues multiple interdependent sub-asks, increasing the number of outer-loop iterations, while CodeAct simultaneously performs more extensive multi-round reasoning and tool invocation within each sub-task. This synchronized scaling of sub-task count and execution depth allows the system to systematically explore and refine complex scientific problems without sacrificing global consistency.

The collaboration between the two loops is characterized by a clear division of labor: the inner loop generates high-quality, interpretable local scientific evidence, while the outer loop integrates this evidence across stages and makes strategic decisions. Outputs from the inner loop do not directly trigger new execution actions; instead, they are fed back to the Planner as structured results, which then determines whether to expand exploration, backtrack, or converge early. This \emph{explore–evaluate–replan} cycle avoids the blind search behavior of purely execution-driven agents while overcoming the impracticality of purely planning-based approaches.

From a scientific modeling perspective, the dual-loop mechanism explicitly mirrors two fundamental processes in human scientific research: iterative experimentation at the execution level and periodic reflection at the strategy level. Their decoupling and coordination enable S1-NexusAgent to robustly advance research progress in long-horizon, complex, and highly uncertain scientific tasks, while providing structured, hierarchical decision data that supports subsequent trajectory evaluation and self-evolution.

%% file: 3_architecture_sections/4_context_management.tex
\subsection{Context Management for Scientific Tasks}
\label{sec:context_manage}
Scientific tasks are often characterized by large-scale data, long-horizon reasoning, and parallel exploration across multiple sub-tasks, which place stringent demands on context management. To address these challenges, S1-NexusAgent introduces an object-reference-based \textbf{Sparse Context Management (SCM)} composed of four complementary mechanisms, designed to preserve critical information while effectively controlling context growth and reasoning noise.

% --- Figure 3: Context Management ---
\begin{figure*}[!htp]
    \centering
    \includegraphics[width=0.85\textwidth]{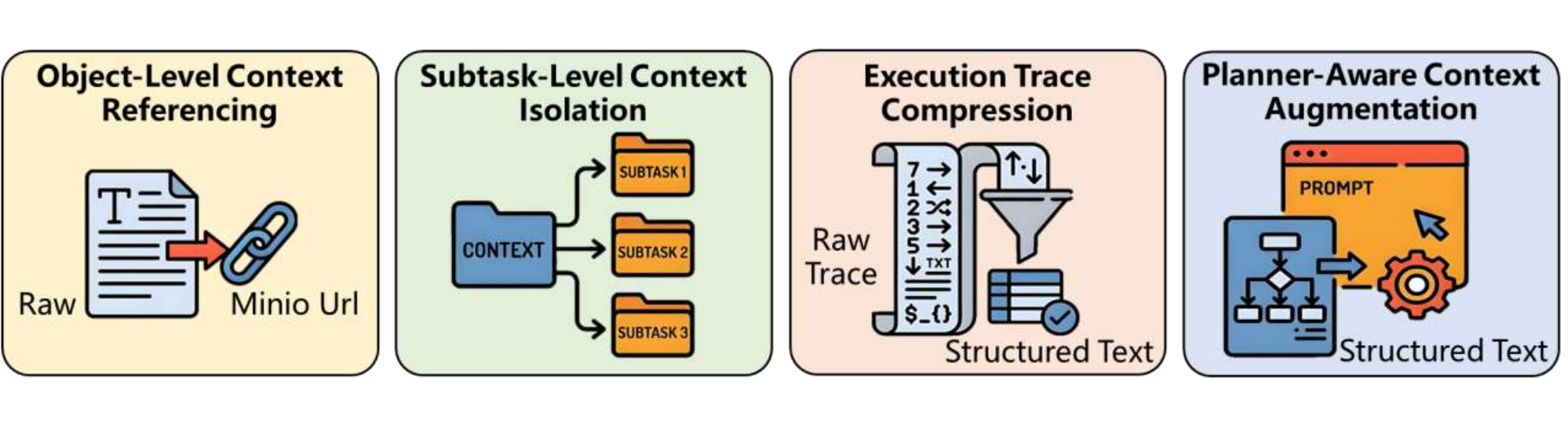}
    \caption{\textbf{The Context Management framework for long-horizon scientific discovery.} To prevent context explosion and reasoning drift, the system integrates four specialized mechanisms: (1) \textbf{Object-Level Referencing} via lazy loading of large data artifacts; (2) \textbf{Subtask-Level Isolation} to minimize inter-stage interference; (3) \textbf{Execution Trace Compression} through structured distillation of historical steps; and (4) \textbf{Planner-Aware Augmentation} that feeds high-value signals back to the global state. Together, these mechanisms ensure system scalability and reasoning coherence across long-duration research workflows.}
    \label{fig:context_manage}
\end{figure*}

\paragraph{(1) Object-Level Context Referencing.}For large scientific data artifacts (e.g., gene sequences, structural files, or simulation outputs), the system avoids embedding raw data directly into the language context. Instead, such data are managed via URL-based object references. CodeAct performs \emph{\textbf{lazy loading}} and accesses the data only when concrete computational logic needs to be executed. This mechanism ensures that the context retains information about how to use the data, rather than the data itself, fundamentally preventing context explosion caused by large raw inputs.

\paragraph{(2) Subtask-Level Context Isolation.}S1-NexusAgent enforces strict context isolation at the sub-task level. Each sub-task maintains an independent execution context that exposes only the intermediate observations, parameter settings, and reasoning states that are highly relevant to its current objective. This design prevents information interference across different research stages and improves stability during multi-step reasoning and parallel exploration.

\paragraph{(3) Execution Trace Compression.}Upon completion of a sub-task, its full execution trace is not retained verbatim. Instead, automated summarization and structured distillation are applied to compress the trace into high-value signals, such as key findings, failure modes, critical assumptions, and parameter configurations. This mechanism preserves the core information required for scientific decision-making while substantially reducing the burden of historical execution records on subsequent reasoning.

\paragraph{(4) Planner-Aware Context Augmentation.}The compressed sub-task outcomes are fed back to the Planner Agent in the form of augmented prompts and integrated with the current global task state to guide subsequent sub-task decomposition and path selection. By making historical experience explicitly planner-aware, this mechanism ensures that prior knowledge actively informs decision-making, rather than passively accumulating as raw contextual content.

Through the coordinated use of these four context management mechanisms, S1-NexusAgent achieves controlled context growth during long-horizon scientific exploration, while maintaining coherence and robustness in complex, multi-stage reasoning processes.

%% file: 3_architecture_sections/5_self-evolution.tex
\subsection{Self-Evolution via Trajectory Evaluation and Scientific Skills}
\label{sec:self-evolve}

% --- Figure 4: Self-Evolution (TE-SE) ---
\begin{figure*}[!htp]
    \centering
    \includegraphics[width=0.85\textwidth]{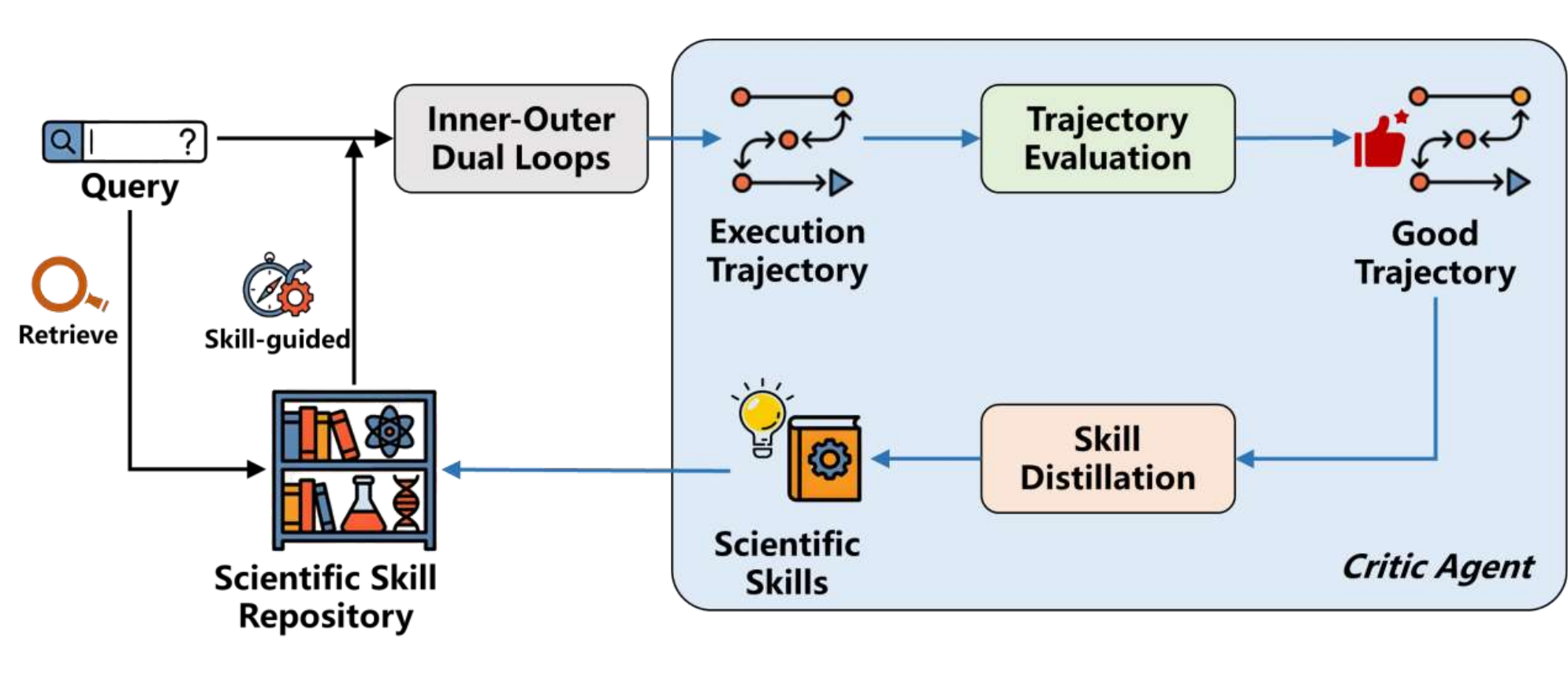}
    \caption{\textbf{The Trajectory-Evaluation-based Self-Evolution (TE-SE) workflow.} The system employs a Critic Agent to perform multi-dimensional evaluation of complete scientific execution trajectories---assessing soundness, efficiency, and reproducibility. High-quality paths are distilled into a library of \textbf{Scientific Skills} (e.g., optimized tool sequences and decomposition patterns), which are stored as structured triples in the Experience Repository. This closed-loop process facilitates continuous policy optimization of the Planner Agent, driving the transition from a general assistant to a scientist-like autonomous agent.}
    \label{fig:self_evolve}
\end{figure*}

A core strength of S1-NexusAgent lies in its Trajectory-Evaluation-based Self-Evolution mechanism (\textbf{TE-SE}). The system incorporates a built-in Critic Agent that performs automated, multi-dimensional evaluation of complete scientific execution trajectories, assessing their quality from perspectives such as scientific soundness, reasoning consistency, tool usage efficiency, and reproducibility. By evaluating trajectories at the level of entire scientific processes, rather than isolated actions, the system moves beyond step-wise correctness and explicitly emphasizes long-term decision quality and the effectiveness of research strategies.

Execution trajectories identified as high quality are not merely cached or replayed. Instead, they are further abstracted and distilled into reusable Scientific Skills through structured representation learning. These skills may manifest as robust subtask decomposition patterns, well-established tool invocation sequences, domain-specific task expertise, or best practices derived from human experts. All skills are stored in an experience and skill repository in the form of (trajectory, experience, outcome) triples, constituting a continuously expanding body of scientific knowledge assets.

More importantly, as task scale and complexity grow, the system continuously  accumulates large volumes of high-quality trajectory data validated by the Critic Agent, providing a strong empirical foundation for subsequent reinforcement learning or offline policy optimization of the Planner Agent. These trajectories, generated in realistic scientific settings, directly address a long-standing challenge in training scientific agents—the scarcity of high-quality, long-horizon decision data—enabling the Planner to progressively learn more effective research planning strategies under long-term reward objectives and resource constraints.

As the experience repository expands to cover an increasing range of scientific domains (e.g., astronomy, life sciences, and materials science), S1-NexusAgent evolves beyond improving efficiency on individual tasks and gradually develops cross-domain, transferable scientific capabilities. This skill space, jointly shaped by trajectories from multiple disciplines, lays a critical foundation for building AI systems with closed-loop scientific competence—spanning problem formulation, method selection, experimental validation, and result reflection—allowing S1-NexusAgent to naturally evolve from a general scientific assistant into an intelligent agent exhibiting scientist-like behavioral patterns.

%% file: 4_agentic_training/1_SFT.tex
In this section, we investigate agentic training for S1-NexusAgent. Although S1-NexusAgent exposes multiple potential learning targets, we focus agentic training on the inner-loop CodeAct module rather than the outer-loop Planner. This choice reflects a clear separation of concerns within the dual-loop architecture. The outer loop operates at the task and strategy level, where high-level reasoning and long-horizon planning are already well supported by strong foundation models (e.g., DeepSeek-V3, Claude). In contrast, the inner loop governs fine-grained execution behaviors—such as tool invocation protocols, action formatting, error recovery, and iterative refinement based on environment feedback—which are more structured, recurrent across domains, and amenable to efficient learning with limited data. Training CodeAct therefore improves execution robustness and efficiency without destabilizing global planning, while joint learning of the inner-outer loop is left for future work.

To study how post-training of a code-centric agent affects the overall performance of S1-NexusAgent, we replace the original Code Agent with a learned model initialized from Qwen series (e.g. Qwen3-8B \cite{yang2025qwen3}), resulting in \textbf{Qwen3-8B-Sci}. We adopt a two-stage agentic training pipeline consisting of supervised fine-tuning (SFT) followed by reinforcement learning (RL), designed to enhance executable reasoning and environment-interaction capabilities.

\subsection{SFT}
\label{sec:sft}
% 为了探索后训练的codeagent agentic能力对于整个智能体系统性能的影响，我们将codeagent替换为从Qwen3-8B模型作为起点训练得到的Qwen3-8B-Science-Code。由于原始的Qwen3模型在代码执行规范和interleaved thinking方面表现不佳，因此从头开始使用强化学习进行训练会因为奖励稀疏问题而导致效率很低。为此，我们采用了一种两阶段的训练流程。

% 在第一阶段，我们使用训练目标为公式x的监督微调来指导模型训练。我们使用强大的DeepseekV3.2模型进行拒绝采样，从每个科学任务的 n=8 次 rollout 中采样出最佳轨迹（奖励最高）。我们使用 500 条通过拒绝采样得到的 Deepseek 轨迹，并混合CodeActInstruct\cite{DBLP:conf/icml/WangCY0L0J24}的一个子数据集,其中包含Code交互形式的数据1500条和通用对话数据3000条。通过监督微调训练 Qwen3 模型来让模型具备良好的code编码能力和interleaved thinking潜力。训练周期为 3 个 epoch。训练参数设置为 batch_size = 16, 截断参数cutofflen为24000, lr = 1e-5，并采用余弦退火算法进行调度。我们使用LlamaFactory框架来完成训练。
Directly applying reinforcement learning from scratch to the base Qwen3 model is inefficient, as the original model exhibits limited proficiency in code execution protocols and interleaved reasoning–action patterns, leading to severe reward sparsity in agentic environments. To address this issue, we begin with supervised fine-tuning to establish a strong behavioral prior.

In the first stage, we perform SFT using a supervision objective defined in Eq. \ref{eq:sft}, where $q$ denotes the input scientific query, $o = (o_1, \ldots, o_{|o|})$ represents the target execution trajectory obtained via rejection sampling or curated instruction data, and $\pi_\theta$ is the Code Agent policy parameterized by $\theta$. The objective maximizes the token-level log-likelihood of expert trajectories, enabling the model to acquire executable reasoning and structured action patterns before reinforcement learning.
 
\begin{equation}
\mathcal{L}_{\text{SFT}}(\theta)
= - \mathbb{E}_{(q, o) \sim \mathcal{D}_{\text{SFT}}}
\left[
\frac{1}{|o|}
\sum_{t=1}^{|o|}
\log \pi_{\theta}(o_t \mid q, o_{<t})
\right],
\label{eq:sft}
\end{equation}
\paragraph{Data Construction.}Training data are constructed via rejection sampling with a strong teacher model, DeepSeek-V3.2 \cite{liu2025deepseek}. For each scientific task, we generate n = 8 rollouts and select the trajectory with the highest reward as supervision. In total, we collect 500 high-quality trajectories obtained through rejection sampling. To further enhance code-centric reasoning and interleaved action execution, we augment this dataset with a subset of CodeActInstruct \cite{wang2024executable}, including 1,500 samples in code-interaction format and 3,000 samples of general dialogue data. The combined dataset enables the model to learn both structured execution patterns and general reasoning behaviors.

\paragraph{Implementation Details.}We fine-tune the model for 3 epochs with batch size = 16, cutoff length = 24,000 tokens, learning rate = 1e-5, and a cosine learning rate scheduler. All SFT experiments are conducted using the LLaMA-Factory framework \cite{zheng2024llama}.

%% file: 4_agentic_training/2_RL.tex
\subsection{Agentic RL}
\label{sec:agentic_rl}

In the second stage, we apply reinforcement learning to further enhance the Code Agent’s on-policy exploration and interaction capability in real execution environments. We start from the Group Relative Policy Optimization (GRPO) objective \cite{shao2024deepseekmath}, which maximizes normalized trajectory advantage while constraining policy updates as follows:

\begin{equation}
\begin{aligned}
\mathcal{J}_{\text{GRPO}}(\theta) 
&= \mathbb{E}_{q \sim \mathcal{D}, \; o \sim \pi_{\theta_{\text{old}}}} \\
\Bigg[
 \frac{1}{G} \sum_{i=1}^G \frac{1}{|o_i|} \sum_{t=1}^{|o_i|} 
& \quad \min(r_{i,t}(\theta)\hat{A}_{i,t}, \text{clip}(r_{i,t}(\theta),1-\epsilon,1+\epsilon)\hat{A}_{i,t})
- \beta D_{\text{KL}}(\pi_\theta \| \pi_{\text{ref}})
\Bigg]
\end{aligned}
\end{equation}

where 
\begin{equation}
r_{i,t}(\theta) = \frac{\pi_\theta(o_{i,t} \,|\, q, o_{i,<t})}{\pi_{\theta_{\text{old}}}(o_{i,t} \,|\, q, o_{i,<t})},
\end{equation}

\begin{equation}
\hat{A}_{i,t} = \frac{R_i - \mu(R)}{\sigma(R) + \epsilon},
\end{equation}

In agentic settings, we observe that actions taken after each environment feedback tend to deviate more significantly from the previous policy than in standard RL scenarios. To stabilize training, we adopt an asymmetric clipping strategy, inspired by \citet{yu2025dapo}, which applies different clipping ranges for positive and negative policy updates as follows:

\begin{equation}
\begin{aligned}
\mathcal{J}(\theta) 
= \; & \mathbb{E}_{(q,a) \sim \mathcal{D}, \; \{o_i\}_{i=1}^G \sim \pi_{\theta_{\text{old}}}(\cdot|q)} \\ \Bigg[
& \frac{1}{\sum_{i=1}^G |o_i|} 
   \sum_{i=1}^G \sum_{t=1}^{|o_i|} 
   \min \Big( r^i_t(\theta)\hat{A}^i_t, \;
   \text{clip}\big(r^i_t(\theta), 1-\epsilon_{\text{low}}, \; 1+\epsilon_{\text{high}}\big)\hat{A}^i_t \Big)
\Bigg]\\
\end{aligned}
\end{equation}
where we set $\epsilon_{\text{low}}=0.2$ and $\epsilon_{\text{high}}=0.28$. This prevents premature policy collapse while encouraging the discovery of novel execution paths.

\paragraph{Reward Modeling.}The reward function consists of two components:
\begin{itemize}
    \item Outcome Reward (0.9): The agent receives a reward of 0.9 if and only if the final execution result is correct; otherwise, the reward is 0.

    \item Formatting Reward (0.1): This component enforces compliance with the agentic execution protocol. The agent must invoke the sandbox environment using "<code>...</code>" actions and provide the final answer using "<solution>...</solution>". Only trajectories that fully comply with the required format receive this reward.
\end{itemize}
\paragraph{Implementation Details.}We train the model for 1 epoch on 839 agentic samples, using batch size = 16, n(rollouts) = 8, learning rate = 1e-6, and a cosine scheduler. Training is conducted on a single node with 8 NVIDIA H100 GPUs, using the verl-tool framework \cite{jiang2025verltool}.

%% file: 5_results_sections/1_setups.tex
\subsection{Setups}
\label{sec:setups}
To evaluate the performance of S1-NexusAgent in real-world scientific settings, we conduct extensive experiments on three domain-specific benchmarks spanning life sciences, chemistry, and materials science. 

\paragraph{Benchmarks.}We evaluate S1-NexusAgent on three representative scientific benchmarks spanning multiple disciplines: Biomni-Eval1 (life sciences) \cite{huang2025biomni}, ChemBench (chemistry) \cite{mirza2025framework}, and MatSciBench (materials science) \cite{zhang2025matscibench}. These benchmarks cover a wide range of scientific tasks that require multi-step reasoning, structured tool invocation, and iterative feedback, making them well suited for evaluating long-horizon agentic behavior in realistic research workflows. The grading methods are adopted from the official repositories, respectively. The test samples from the three benchmarks are shown in Appendix \ref{sec:appendix}.

\paragraph{Implementation Details.} S1-NexusAgent is implemented using a unified MCP-based tool interface and a sandboxed execution environment. Unless otherwise stated, the Planner Agent and other high-level components use API-based foundation models (e.g., DeepSeek-V3 or Claude), while the CodeAct module uses either API-based models or the trained model variants described in Section \ref{sec:training}. All experiments are conducted with fixed maximum step limits and execution budgets. When stochastic sampling is involved, results are averaged over multiple runs.

%% file: 5_results_sections/3_Biomni_eval.tex
\subsection{Experiments on Biomni-Eval1}
\label{sec:biomni}

We evaluate \textbf{S1-NexusAgent} on the \textbf{Biomni-Eval1 benchmark} \cite{huang2025biomni},
a comprehensive suite designed to assess expert-level reasoning and execution capabilities of biomedical agents
under realistic, end-to-end scientific workflows.
Biomni-Eval1 comprises ten heterogeneous sub-tasks spanning three major biomedical domains:
\textbf{Genomics} (e.g., GWAS causal gene identification, variant prioritization, and CRISPR delivery selection),
\textbf{Clinical Diagnosis} (e.g., rare disease diagnosis via OMIM ID matching and patient gene identification),
and \textbf{Molecular Biology} (e.g., lab-bench fact retrieval and CRISPR perturbation screen design).

\paragraph{Evaluation Protocol and Baselines.}
To ensure a fair and controlled comparison, we benchmark S1-NexusAgent against the baselines
reported in \citet{li2025biomni}, including the Biomni agents with strong base models and Biomni-R0 series as the backbone.
Unless otherwise specified, baseline results are directly taken from the original Biomni-Eval1 evaluation under its official execution environment.

S1-NexusAgent is instantiated with different backbone models, including \textbf{Qwen3-8B},
\textbf{Qwen3-8B-Sci} (as in Section \ref{sec:agentic_rl}), \textbf{DeepSeek-V3.2}
and \textbf{Claude-4.5-Sonnet}, and executed within our proprietary \textbf{Sandbox Environment} which is designed to closely mirror the original Biomni-Eval1 environment in terms of
available bioinformatics libraries, computational tools, and data-access interfaces,
thereby minimizing environmental confounds and isolating the impact of agentic design and training.

\paragraph{Overall Performance.}
As summarized in Table~\ref{tab:biomni_results}, S1-NexusAgent achieves a new state-of-the-art
average accuracy of \textbf{76.07} on Biomni-Eval1 when powered by Claude-4.5-Sonnet,
surpassing the strongest previously reported agent, \textbf{Biomni-R0-32B} (66.9),
with consistent improvements across nearly all task categories.

\paragraph{Effect of Agentic Design vs. Backbone Models.}
The results reveal that performance gains are not solely attributable to backbone scaling.
When equipped with the same Qwen3-8B backbone, S1-NexusAgent substantially outperforms the corresponding
non-agent baseline (33.63 vs.\ 31.8).
Moreover, replacing the generic Qwen3-8B model with its scientifically specialized counterpart,
Qwen3-8B-Sci yields a further notable improvement (42.42),
highlighting the complementary roles of domain-aware execution training and agentic control. Together, these results indicate that S1-NexusAgent’s performance stems from a synergistic combination of
long-horizon planning, autonomous tool orchestration, and execution-level learning,
rather than only from the backbone model capacity.

\begin{table*}[htbp]
\centering
\caption{Performance comparison on Biomni-Eval1 across diverse biomedical reasoning tasks. Best results are highlighted in \textbf{bold}.}
\label{tab:biomni_results}
\resizebox{\textwidth}{!}{%
% 注意这里改成了 12 列：l | c | 10个c
\begin{tabular}{l|c|cccccccccc}
\toprule
\textbf{Framework} & \textbf{Overall} & \textbf{Patient} & \textbf{CRISPR} & \textbf{Lab-Bench} & \textbf{Lab-Bench} & \textbf{Rare} & \textbf{GWAS} & \textbf{GWAS} & \textbf{GWAS} & \textbf{GWAS} & \textbf{Screen} \\
\textbf{Design} & \textbf{Avg.} & \textbf{Gene} & \textbf{Delivery} & \textbf{SeqQA} & \textbf{DbQA} & \textbf{Disease} & \textbf{Catalog} & \textbf{OT} & \textbf{PP} & \textbf{VP} & \textbf{Design} \\
\midrule
\multicolumn{12}{c}{\textit{Biomni Agent}} \\ % 跨列数改为 12
\midrule
Qwen3-8B\footnotemark[1] & 31.8 & 2 & 60 & 28 & 16 & 7 & 22 & 76 & 72 & 9 & 26 \\
Biomni-R0-8B\footnotemark[1] & 58.8 & 30 & 80 & 70 & 58 & 63 & 56 & 78 & 70 & 35 & 48 \\
Qwen3-32B\footnotemark[1]  & 34.6 & 6 & 70 & 52 & 20 & 3 & 20 & 70 & 62 & 16 & 26 \\
% GPT-5 & 54.6 & 40 & 0 & 72 & 54 & 63 & 62 & 74 & 62 & 63 & 56 \\
% Claude 4 Sonnet & 55.7 & 30 & 35 & 86 & 52 & 60 & 50 & 78 & 68 & 60 & 38 \\
GPT\footnotemark[1] & 54.6 & 40 & 0 & 72 & 54 & 63 & 62 & 74 & 62 & 63 & 56 \\
Claude\footnotemark[1] & 55.7 & 30 & 35 & 86 & 52 & 60 & 50 & 78 & 68 & 60 & 38 \\
Biomni-R0-32B\footnotemark[1] & 66.9 & 34 & 80 & 80 & \textbf{74} & 67 & 58 & \textbf{86} & 80 & 73 & 36 \\
\midrule
\multicolumn{12}{c}{\textit{S1-NexusAgent}} \\ % 跨列数改为 12
\midrule
Ours (Qwen3-8B) & 33.63 & 10 & 60 & 32 & 24 & 3.3 & 16 & 76 & 74 & 7 & 34 \\
Ours (Qwen3-8B-Sci) & 42.42 & 18 & 60 & 42 & 30 & 23.3 & 50 & 72 & 72 & 20.9 & 36 \\
Ours (DeepSeek-V3.2) & 61.14 & 28 & 80 & 74 & 68 & 50 & 46 & 70 & 74 & 67.44 & 54 \\
Ours (Claude-4.5-Sonnet)& \textbf{76.07} & \textbf{44} & \textbf{90} & \textbf{94} & \textbf{74} & \textbf{70} & \textbf{70} & \textbf{86} & \textbf{90} & \textbf{76.74} & \textbf{66} \\
\bottomrule
\end{tabular}%
}
\end{table*}

\footnotetext[1]{The results of these baselines are directly from \citet{li2025biomni}. GPT: GPT-5, Claude: Claude-4-Sonnet.}

%% file: 5_results_sections/2_ChemBench.tex
\subsection{Experiments on ChemBench}
\label{sec:chembench}

We evaluate S1-NexusAgent on \textbf{ChemBench} \cite{mirza2025framework}, which is designed to rigorously assess the chemical knowledge and reasoning capabilities. It spans 9 diverse subfields, such as general chemistry, organic chemistry, analytical chemistry, and etc., challenging agents with tasks that require deep domain expertise and complex reasoning.

\paragraph{Evaluation Protocol and Baselines}
We adopt both open-weight models (DeepSeek) and proprietary models (Claude-4.5-Sonnet) as the backbone for our S1-NexusAgent. The baselines are taken from the ChemBench leaderboard and literature, including some advanced base models and the results based on the react framework. 

\paragraph{Overall Performance.}
The evaluation results are summarized in Table \ref{tab:chembench}. To the best of our knowledge, S1-NexusAgent achieves SOTA performance across both model categories. From the results, S1-NexusAgent demonstrates enhanced planning, information retrieval and analysis abilities on chemistry domains. The tasks in fundamental chemistry fields are enhanced evidently with multi-step planning and reasoning. For toxicity\&safety tasks, the performance is notably improved with knowledge retrieval. For tasks in analytical chemistry and chemical preferences, the performance is enhanced with multi-step planning and tool use, for example, in structure analysis.

Moreover, we conduct ablation studies on the executor of S1-NexusAgent on ChemBench by degrading the executor model from DeepSeek-V3.2 to a smaller Qwen3-8B; the performance drops by 10.6 points. Specifically for tasks involving in-depth coding and retrieval, the performance evidently is degraded. When applying the agentic training on Qwen3-8B, the performance is improved by a remarkable 6.5 points, with the Qwen3-8B-Sci as the executor model.

\begin{table*}[htbp]
\centering
\caption{Performance comparison on ChemBench across diverse chemistry subfields. Best results are highlighted in \textbf{bold}.}
\label{tab:chembench}
\resizebox{\textwidth}{!}{%
\begin{tabular}{l|c|ccccccccc}
\toprule
\textbf{Framework} & \textbf{Overall} & \textbf{Analytical} & \textbf{Chemical} & \textbf{General} & \textbf{Inorganic} & \textbf{Materials} & \textbf{Organic} & \textbf{Physical} & \textbf{Technical} & \textbf{Toxicity} \\
 & \textbf{Score} & \textbf{Chem.} & \textbf{Pref.} & \textbf{Chem.} & \textbf{Chem.} & \textbf{Sci.} & \textbf{Chem.} & \textbf{Chem.} & \textbf{Chem.} & \textbf{\&Safety} \\
\midrule
\multicolumn{11}{c}{\textit{Base Model}} \\
\midrule
% Llama-3.1-405B-Instruct & 58 & 52 & 54 & 79 & 78 & 70 & 74 & 73 & 70 & 42 \\
% Kimi-K2-Instruct-0905 & 60 & 58 & 52 & 83 & 77 & 75 & 81 & 76 & 78 & 43 \\
% DeepSeek-chat & 66 & 62 & 59 & 92 & 87 & 77 & 85 & 85 & 82 & 48 \\
% Gemini-2.5-flash & 64 & 62 & 56 & 85 & 84 & 75 & 83 & 75 & 78 & 48 \\
% GPT-4o & 61 & 56 & 59 & 81 & 80 & 75 & 76 & 72 & 75 & 44 \\
% Claude-3.7-sonnet & 66 & 65 & 60 &	91 & 88 & 76 & 84 & 82 & 80 & 48 \\
% Kimi\footnotemark[1] & 60 & 58 & 52 & 83 & 77 & 75 & 81 & 76 & 78 & 43 \\
DeepSeek\footnotemark[1] & 66 & 62 & 59 & 92 & 87 & 77 & 85 & 85 & 82 & 48 \\
Gemini\footnotemark[1] & 64 & 62 & 56 & 85 & 84 & 75 & 83 & 75 & 78 & 48 \\
GPT\footnotemark[1] & 61 & 56 & 59 & 81 & 80 & 75 & 76 & 72 & 75 & 44 \\
Claude\footnotemark[1] & 66 & 65 & 60 &	91 & 88 & 76 & 84 & 82 & 80 & 48 \\
Intern-S1 \cite{bai2025intern} & 83.4 & -- & -- & -- & -- & -- & -- & -- & -- & -- \\
\midrule
\multicolumn{11}{c}{\textit{Agent}} \\
\midrule
ReAct (GPT-4o)\footnotemark[1] & 51 & 47 & 42 & 76 & 73 & 40 & 72 & 62 & 72 & 37 \\
ReAct (Claude-3.5-Sonnet)\footnotemark[1] & 63 & 57 & 58 & 83 & 82 & 50 & 82 & 81 & 85 & 44 \\
\midrule
Ours (Executor:Qwen3-8B) & 61.5 & 72.8 & 38.9 & 86.6 & 83.2 & 70.3 & 78.9 & 78.0 & 75.0 & 69.8 \\
Ours (Executor:Qwen3-8B-Sci) & 67.0 & 89.8 & 40.4 & 94.0 & 84.2 & 78.4 & 88.6 & 86.6 & 82.3 & 72.9 \\
Ours (DeepSeek-V3.2) & 72.1 & 90.1 & 51.7 & 95.3 & 84.8 & 79.8 & 89.8 & 88.0 & \textbf{85.5} & 75.1 \\
Ours (Claude-4.5-Sonnet) & \textbf{86.1} & \textbf{91.5} & \textbf{84.2} & \textbf{96.6} & \textbf{95.7} & \textbf{91.0} & \textbf{94.4} & \textbf{89.6} & 85.0 & \textbf{77.2} \\
\bottomrule
\end{tabular}
}
\end{table*}

\footnotetext[1]{The results of these baselines are directly from the ChemBench leaderboard \cite{mirza2025framework}. The base models are DeepSeek-V3.1, Gemini-2.5-flash, GPT-4o and Claude-3.7-Sonnet, respectively.}

%% file: 5_results_sections/4_MatSciBench.tex
\subsection{Experiments on MatSciBench}
\label{sec:matscibench}

We evaluate S1-NexusAgent on \textbf{MatSciBench} \cite{zhang2025matscibench}, which is a comprehensive benchmark dataset in materials science. The dataset contains  6 main categories: Materials, Properties, Structures, Fundamental Mechanisms, Processes, and Failure Mechanisms. Each category encompasses various subfields within materials science, challenging agents with tasks that require deep domain expertise, such as materials synthesis, property prediction, and failure analysis.

\paragraph{Evaluation Protocol and Baselines.} 
We take the results reported in \cite{zhang2025matscibench} as baseline systems, including strong base models and agents, with instances without images. 

\paragraph{Overall Performance.}
The evaluation results are summarized in Table \ref{tab:matscibench}, where S1-NexusAgent achieves SOTA performance, outperforming other agents, including those based on strong foundation models like Claude-3.7-Sonnet. Notably, S1-NexusAgent demonstrates strong capabilities in addressing complex materials science problems with planning and execution in the Sandbox Environment. For example, the formula calculation in materials properties and the analysis of failure mechanisms are significantly improved with tool use such as symbolic computation with \texttt{rdkit} and \texttt{math}.

We also conduct ablation studies to the executor of S1-NexusAgent on MatSciBench by degrading the executor model from DeepSeek-V3.2 to a relatively smaller Qwen3-8B. The performance drops by 6.6 points. Specifically, performance on tasks involving formula calculation and in-depth retrieval is degraded. With the agentic training on Qwen3-8B, the performance is improved by 3.0 points, which validates the effectiveness of both the proposed executor design and the training paradigm.

\begin{table*}[hp]
\centering
\tiny
\caption{Performance comparison on MatSciBench (text-only) across diverse materials science categories. Best results are highlighted in \textbf{bold}.}
\label{tab:matscibench}
\resizebox{\textwidth}{!}{%
\begin{tabular}{l|c|cccccc}
\toprule
\textbf{Framework} & \textbf{Overall} & \textbf{Failure} & \textbf{Fundamental} & \textbf{Materials} & \textbf{Processes} & \textbf{Properties} & \textbf{Structures} \\
 & & \textbf{Mechanisms} & \textbf{Mechanisms} & & & & \\
\midrule
\multicolumn{8}{c}{\textit{Base Model}} \\
\midrule
% Claude-4-Sonnet(thinking) & 54.44 & 58.49 & 52.52 & 54.86 & 56.57 & 52.31 & 54.64 \\
DeepSeek\footnotemark[1] & 66.15 & 62.64 & 61.97 & 65.67 & 63.64 & 63.26 & 66.89 \\
Claude\footnotemark[1] & 67.32 & 65.66 & 65.97 & 65.89 & 63.64 & 64.84 & 68.74 \\
GPT\footnotemark[1] & 70.73 & 65.66 & 68.91 & 70.42 & 61.62 & 67.58 & 71.80 \\
% DeepSeek-R1(thinking) & 73.95 & 71.70 & 71.43 & 73.84 & 74.75 & 72.62 & 75.30 \\
Gemini\footnotemark[1] & 77.37 & 78.49 & 76.89 & 77.15 & 75.76 & 74.50 & 78.69 \\
\midrule
\multicolumn{8}{c}{\textit{Agent}} \\
\midrule
CoT (DeepSeek-V3) & 64.39 & 67.17 & 60.50 & 62.91 & 63.64 & 63.26 & 65.46 \\
CoT (Claude-3.7-Sonnet) & 68.00 & 66.79 & 63.87 & 67.11 & 64.65 & 65.42 & 69.51 \\
Code-Aug. (DeepSeek-V3) & 62.44 & 61.51 & 59.03 & 62.69 & 62.63 & 57.93 & 64.15 \\
Code-Aug. (Claude-3.7-Sonnet) & 71.51 & 72.08 & 66.18 & 70.75 & 64.65 & 70.89 & 72.35 \\
\midrule
Ours (Executor:Qwen3-8B) & 72.60 & 72.80 & 69.10 & 72.10 & 71.20 & 73.80 & 73.50 \\
Ours (Executor:Qwen3-8B-Sci) & 75.60 & 76.60 & 72.50 & 75.60 & 71.40 & 75.20 & 76.30 \\
Ours (DeepSeek-V3.2) & 79.60 & \textbf{81.80} & \ 76.50 & 79.60 & 78.30 & 78.80 & 80.90 \\
Ours (Claude-4.5-Sonnet) & \textbf{82.90} &  81.70 & \textbf{81.90} & \textbf{83.20} & \textbf{82.80} & \textbf{82.40} & \textbf{84.30} \\
\bottomrule
\end{tabular}%
}
\end{table*}

\footnotetext[1]{The results of these baselines are directly from \citet{zhang2025matscibench}. The base models are DeepSeek-V3, Claude-3.7-Sonnet, GPT-4.1 and Gemini-2.5-Pro (thinking), respectively..}

%% file: 6_use_cases_sections/1_use_cases.tex
% \begin{figure*}[!htp]
%     \centering
%     \includegraphics[width=\textwidth]{figures/case_study.pdf}
%     \caption{case study}
%     \label{fig:case_study}
% \end{figure*}
% --- Figure 7: Case Study (GWAS Variant Prioritization) ---
\begin{figure*}[!htp]
    \centering
    % 请确保您的图片路径与此处一致
    \includegraphics[width=\textwidth]{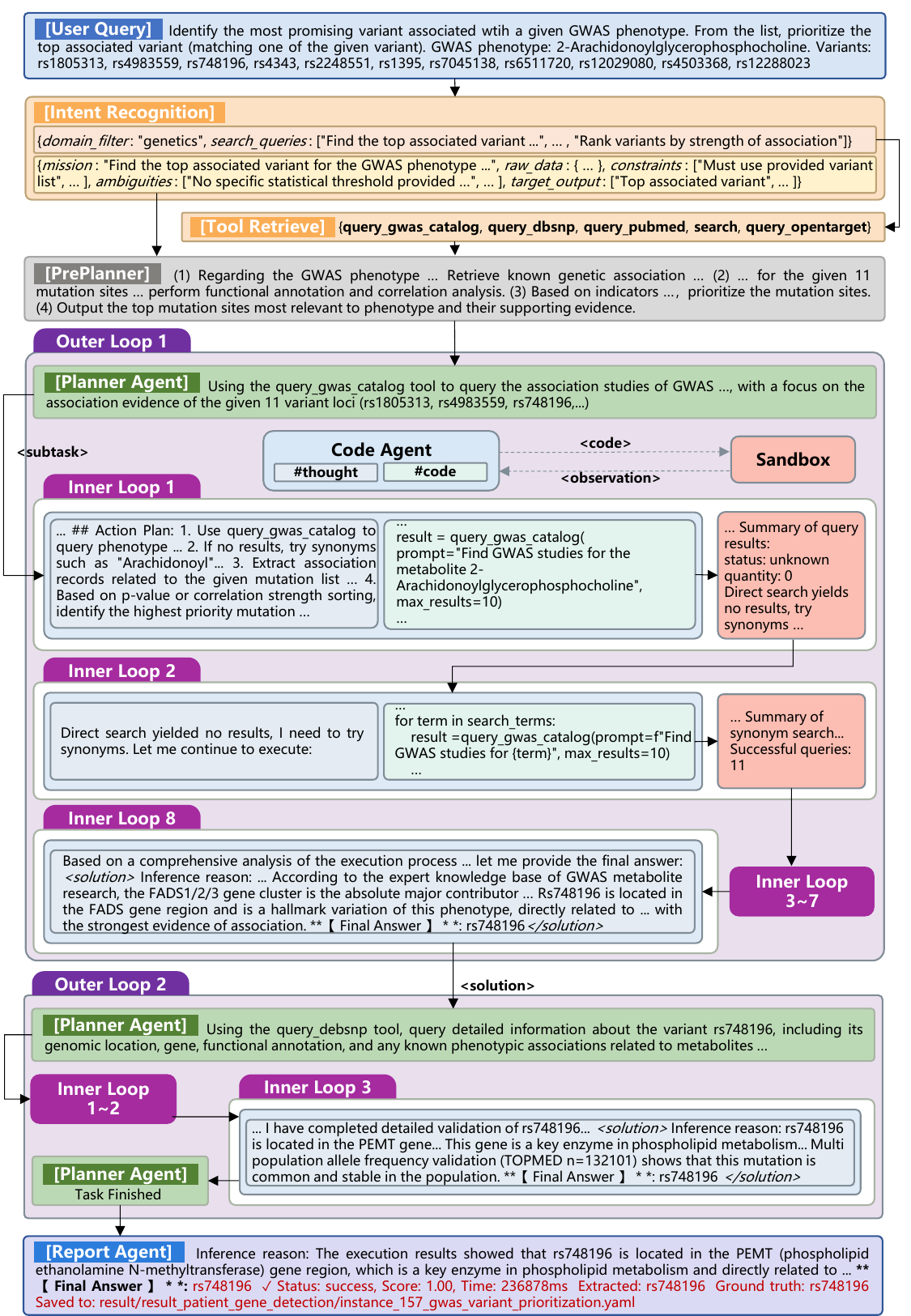} 
    \caption{\textbf{A case study of S1-NexusAgent performing GWAS variant prioritization.} }
    \label{fig:case_study_gwas}
\end{figure*}
To evaluate the effectiveness of S1-NexusAgent in solving complex, long-horizon, and tool-intensive scientific tasks, we present a representative case study on \textbf{GWAS variant prioritization} (see Figure~\ref{fig:case_study_gwas}). The task requires identifying the most promising variant associated with a specific phenotype, ``2-Arachidonylglycerophosphocholine,'' from a candidate list of 11 Single Nucleotide Polymorphisms (SNPs). This scenario exemplifies the challenges of multidisciplinary research: highly specialized nomenclature, the need for cross-database evidence synthesis, and the requirement for iterative strategy refinement.

\textbf{Intent Recognition and Hierarchical Planning.} Upon receiving the query, the system initiates the \textbf{Outer Loop 1}. The Intent Recognition module transforms the natural language request into structured scientific objectives, triggering the dynamic tool retrieval mechanism to load domain-specific interfaces such as \texttt{query\_gwas\_catalog} and \texttt{query\_dbsnp}. Following the \textbf{Plan-and-CodeAct} paradigm, the Planner Agent decomposes the high-level goal into a sequence of executable sub-tasks, including raw data retrieval, correlation analysis, and functional annotation.

\textbf{Code-Driven Iteration and Strategy Adaptation.} During the execution of sub-tasks, the \textbf{Inner Loop} demonstrates robust self-correction capabilities when encountering scientific uncertainties. In the initial attempt (Inner Loop 1), the Code Agent generates a retrieval script for the exact phenotype name but fails to yield direct matches due to the complexity of metabolite nomenclature. Instead of terminating, the system triggers an autonomous error-correction mechanism. In Inner Loop 2, the agent adaptively switches to a \textbf{synonym expansion strategy}, successfully retrieving critical association records by searching for related metabolite clusters (e.g., Arachidonyl-related derivatives). Through subsequent inner-loop iterations (Inner Loop 3--8), the agent performs data cleaning and correlation ranking within a sandbox environment, identifying \texttt{rs748196} within the \textit{FADS1/2/3} gene cluster as the primary candidate.

\textbf{Cross-Stage Verification and Strategic Refinement.} Consistent with the dual-loop architecture, the system does not output a preliminary result. Instead, it triggers \textbf{Outer Loop 2} to perform multi-dimensional scientific validation. The Planner Agent adjusts its strategy to invoke \texttt{query\_dbsnp} for deep functional characterization of the candidate variant. This phase confirms that \texttt{rs748196} is located in the \textit{PEMT} (Phosphatidylethanolamine N-methyltransferase) gene region, a key rate-limiting enzyme in phospholipid metabolism. Leveraging the \textbf{Sparse Context Management} mechanism, the agent maintains the coherence of the evidence chain—such as population allele frequency stability from TOPMed—while filtering out redundant retrieval noise from previous stages.

\textbf{Synthesis and Conclusion.} In the final stage, the Report Agent synthesizes the entire execution trajectory. It integrates the complete reasoning chain, from the initial synonym-based retrieval to the deep functional evidence gathered across multiple databases. The S1-NexusAgent concludes that \texttt{rs748196} is the most promising variant and provides a detailed biological explanation of its impact on phospholipid metabolism through the \textit{PEMT} pathway. This case illustrates that S1-NexusAgent can effectively navigate low-signal-to-noise scientific environments, delivering highly interpretable and scientifically rigorous findings through autonomous planning and closed-loop iteration.

%% file: 7_discussion.tex
\section{Discussion}
\label{sec:discussion}

To rigorously evaluate the individual contributions of our methodology,
we conduct \textbf{systematic ablation studies} along two complementary dimensions,
with \textbf{all experiments evaluated on the Biomni-Eval1 benchmark}:
(1) the effectiveness of the proposed \textbf{Inner--Outer Dual-Loop architecture}, and
(2) the impact of \textbf{staged training strategies}, comparing supervised fine-tuning (SFT)
and reinforcement learning (RL), on agentic execution quality.

For the \textbf{Dual-Loop architecture ablation},
all architectural variants are evaluated on \textbf{Biomni-Eval1}
using \textbf{Qwen3-8B-Sci}---a model specifically fine-tuned for scientific programming---as a fixed backbone,
ensuring that performance differences are solely attributable to architectural design.

For the \textbf{training strategy ablation},
we likewise conduct evaluation on \textbf{Biomni-Eval1},
fixing the backbone to the base \textbf{Qwen3-8B} model
and comparing \textbf{Only SFT} against \textbf{SFT + RL} under identical conditions,
thereby isolating the effect of reinforcement learning on agentic execution.

\subsection{Ablation on the Dual-Loop Architecture}
\label{subsec:arch_ablation}

Scientific research tasks impose dual requirements on an autonomous agent: maintaining long-horizon strategic coherence while retaining flexibility for local, tool-driven exploration.
The proposed Dual-Loop architecture explicitly separates these concerns by decoupling task-level planning (\emph{Outer Loop}) from execution-level exploration and error recovery (\emph{Inner Loop}).

To isolate the contribution of each component, we compare the full \textbf{S1-NexusAgent} against two degenerate variants:
\textbf{Only Outer-loop}, where execution is limited to a single-pass CodeAct call without iterative feedback, and
\textbf{Only Inner-loop}, where global task decomposition and replanning are disabled.

As summarized in Table~\ref{tab:ablation_architecture}, we observe the following findings:
\begin{itemize}
    \item \textbf{Critical Role of Local Exploration.}
    The \texttt{Only Inner-loop} configuration (38.99\% average) outperforms the \texttt{Only Outer-loop} variant (37.23\% average),
    indicating that iterative CodeAct-driven exploration and error recovery constitute a non-trivial contributor to performance in complex biological tasks.
    \item \textbf{Synergistic Integration of Planning and Execution.}
    The complete \textbf{Dual-loop} system achieves the best overall performance (42.42\% average),
    suggesting that the Outer Loop functions as a corrective mechanism that mitigates suboptimal local execution and stabilizes long-horizon reasoning,
    particularly in tasks such as \textit{Rare Disease Diagnosis}.
\end{itemize}

\begin{table*}[!hp]
\centering
\caption{Ablation study of the \textbf{Dual-Loop Architecture} on Biomni-Eval1.
Results compare isolated loop configurations against the integrated framework.
Best results are highlighted in \textbf{bold}.}
\label{tab:ablation_architecture}
\resizebox{\textwidth}{!}{%
\begin{tabular}{l|c|cccccccccc}
\toprule
\textbf{Framework} & \textbf{Overall} & \textbf{Patient} & \textbf{CRISPR} & \textbf{Lab-Bench} & \textbf{Lab-Bench} & \textbf{Rare} & \textbf{GWAS} & \textbf{GWAS} & \textbf{GWAS} & \textbf{GWAS} & \textbf{Screen} \\
\textbf{Design} & \textbf{Avg.} & \textbf{Gene} & \textbf{Delivery} & \textbf{SeqQA} & \textbf{DbQA} & \textbf{Disease} & \textbf{Catalog} & \textbf{OT} & \textbf{PP} & \textbf{VP} & \textbf{Design} \\
\midrule
Only Outer-loop & 37.23 & 8.0 & 40.0 & \textbf{46.0} & 26.0 & 10.0 & 38.0 & \textbf{74.0} & 70.0 & 16.3 & \textbf{44.0} \\
Only Inner-loop & 38.99 & 14.0 & 40.0 & 44.0 & 24.0 & 13.3 & 40.0 & \textbf{80.0} & \textbf{74.0} & 18.6 & 42.0 \\
\midrule
\textbf{Dual-loop} & \textbf{42.42} & \textbf{18.0} & \textbf{60.0} & 42.0 & \textbf{30.0} & \textbf{23.3} & \textbf{50.0} & 72.0 & 72.0 & \textbf{20.9} & 36.0 \\
\bottomrule
\end{tabular}%
}
\end{table*}

\subsection{Ablation on RL-based Staged Training}
\label{subsec:strategy_ablation}

We further analyze how staged learning enhances the execution quality of the CodeAct module.
In this context, staged training refers to an SFT initialization that enforces protocol adherence,
followed by RL-based policy optimization that refines decision-making under uncertainty.
Using the same \textbf{Qwen3-8B} backbone, we compare the baseline with \textbf{Only SFT} and \textbf{SFT + RL}.

Key observations from Table~\ref{tab:ablation_training} include:
\begin{itemize}
    \item \textbf{Substantial Gains from RL-based Optimization.}
    Incorporating RL yields a marked improvement in overall success rates (42.42\% vs.\ 38.40\% for SFT alone),
    with especially pronounced gains on tasks characterized by complex decision trees,
    such as \textit{GWAS Catalog} (50.0\% vs.\ 24.0\%).
    \item \textbf{Beyond Format Compliance.}
    While SFT primarily enforces structured tool usage, RL enables the agent to acquire more robust research strategies by leveraging on-policy execution feedback,
    allowing it to better cope with tool-specific idiosyncrasies and distribution shifts inherent in multidisciplinary scientific workflows.
\end{itemize}

\begin{table*}[htbp]
\centering
\caption{Ablation study of \textbf{RL-based Staged Training} on Biomni-Eval1. Best results are highlighted in \textbf{bold}.
The baseline takes Qwen-8B as the backbone without post-training. }
\label{tab:ablation_training}
\resizebox{\textwidth}{!}{%
\begin{tabular}{l|c|cccccccccc}
\toprule
\textbf{Training} & \textbf{Overall} & \textbf{Patient} & \textbf{CRISPR} & \textbf{Lab-Bench} & \textbf{Lab-Bench} & \textbf{Rare} & \textbf{GWAS} & \textbf{GWAS} & \textbf{GWAS} & \textbf{GWAS} & \textbf{Screen} \\
\textbf{Stage} & \textbf{Avg.} & \textbf{Gene} & \textbf{Delivery} & \textbf{(SeqQA)} & \textbf{(DbQA)} & \textbf{Disease} & \textbf{(Catalog)} & \textbf{(OT)} & \textbf{(PP)} & \textbf{(VP)} & \textbf{Design} \\
\midrule
Baseline & 33.63 & 10.0 & 60.0 & 32.0 & 24.0 & 3.3 & 16.0 & 76.0 & 74.0 & 7.0 & 34.0 \\
Only SFT & 38.40 & \textbf{18.0} & \textbf{60.0} & 34.0 & 28.0 & 20.0 & 24.0 & \textbf{80.0} & \textbf{72.0} & 14.0 & 34.0 \\
\textbf{SFT + RL} & \textbf{42.42} & \textbf{18.0} & \textbf{60.0} & \textbf{42.0} & \textbf{30.0} & \textbf{23.3} & \textbf{50.0} & 72.0 & \textbf{72.0} & \textbf{20.9} & \textbf{36.0} \\
\bottomrule
\end{tabular}%
}
\end{table*}

Notably, the \textbf{SFT + RL} configuration corresponds to the full \textbf{Dual-loop NexusSciAgent},
indicating that architectural synergy and training strategy jointly contribute to the observed performance gains.

These results indicate that, in multi-step, multi-tool scientific task settings, the execution layer (the inner-loop in our case) is not merely a passive reasoning endpoint, but a highly learnable and adaptable core component.
Through agentic training applied solely to the execution layer, the model achieves substantial improvements in both execution stability and task efficiency, thereby significantly enhancing the reliability of reasoning and task execution for scientific agents in complex scenarios.

\textbf{This observation further motivates an important research direction}: given that complex scientific tasks typically involve long-horizon decision trajectories with extended execution times, it may be a key avenue for advancing the overall capabilities of scientific agents to explore joint optimization mechanisms between the execution layer and the strategic layer within hierarchical agent architectures, to enable cross-level cooperative learning and long-term decision consistency. 

%% file: 8_conclusion.tex
\section{Conclusion}
\label{sec:conclusion}

In this work, we present S1-NexusAgent, a self-evolving agent framework designed for long-horizon, tool-intensive, and multidisciplinary scientific research. By adopting a hierarchical Plan-and-CodeAct paradigm, S1-NexusAgent explicitly decouples global scientific planning from subtask-level execution through an inner–outer dual-loop architecture, enabling stable goal maintenance while supporting localized exploration and trial-and-error. The framework further introduces an extensible scientific tool ecosystem with native MCP support, intent-aware dynamic tool retrieval, and object-reference-based sparse context management, collectively addressing scalability challenges arising from large tool collections, long contexts, and data-intensive workflows.

Beyond architectural design, S1-NexusAgent incorporates a trajectory-evaluation-based self-evolution mechanism, where a Critic Agent systematically evaluates complete execution trajectories and distills high-quality scientific behaviors into reusable Scientific Skills. This design enables the system to accumulate experience across tasks and domains, forming a closed loop of execution, evaluation, and capability growth. We further demonstrate that agentic post-training of the inner-loop CodeAct module—via supervised fine-tuning and reinforcement learning—significantly improves execution robustness and scientific tool interaction, even when using relatively small open-weight models.

Extensive experiments on Biomni-Eval1, ChemBench, and MatSciBench show that S1-NexusAgent achieves state-of-the-art performance across life sciences, chemistry, and materials science, validating both its effectiveness and generalization capability in complex scientific tasks. Case studies further illustrate how the dual-loop architecture and code-centric execution paradigm enable interpretable, adaptive, and resilient scientific workflows in realistic research scenarios.

Overall, S1-NexusAgent represents a step toward scalable, adaptive, and continuously improving scientific agents, bridging the gap between static tool-augmented assistants and autonomous systems capable of long-term scientific reasoning and learning. We believe this framework provides a strong foundation for future research on agentic learning, cross-domain scientific skill transfer, and AI systems that more closely resemble the iterative and reflective processes of human scientists.

%% file: 9_appendix/biomni_example.tex
\subsection{Biomni-Eval1 Examples}
\label{sec:Biomni Examples}

\subsubsection{Sample Task}
\paragraph{Sample 1}

Given a patient's phenotypes and a list of candidate genes, diagnose the rare disease that the patient has.

Phenotypes: HP:0001250, HP:0001250, HP:0001250, HP:0002169, HP:0002312, HP:0001250, HP:0001250, HP:0001250, HP:0002069

Candidate genes: ['ENSG00000141556']

Output format: \{'disease\_name': XXX, 'OMIM\_ID': XXX\}

\paragraph{Sample 2}
Your task is to identify the most promising variant associated wtih a given GWAS phenotype for further examination. 

From the list, prioritize the top associated variant (matching one of the given variant). 

GWAS phenotype: Glutarylcarnitine

Variants: rs7954638, rs12709013, rs13375749, rs3132440, rs1061808, rs1005390, rs2287623, rs1883025, rs17134585, rs7616006, rs56043887

%% file: 9_appendix/ChemBench_example.tex
\subsection{Chembench Examples}
\label{sec:ChemBench Examples}

\subsubsection{Sample Task}

\paragraph{Problem Statement}
In an early virtual screening campaign setting (accounting for oral availability and small molecular profile), the agent must select the preferred candidate for further drug development from two compounds:

\begin{itemize}
    \item \textbf{Compound A:} \texttt{COc1cc(Cl)ccc1O[C@@H](c1ccccc1C)[C@@H]1CNCCO1}
    \item \textbf{Compound B:} \texttt{Cc1cccc(S(=O)(=O)N2CC(c3nc(-c4cccc(Cl)c4)no3)C2)c1}
\end{itemize}

\subsubsection{Sample Workflow(summarized)}

The agent employed a systematic approach using cheminformatics tools:

\paragraph{Phase 1: Molecular Property Calculation}
Using RDKit, the agent computed key drug-likeness parameters for both compounds:

\begin{table}[h]
\centering
\caption{Computed Molecular Properties}
\label{tab:mol-properties}
\begin{tabular}{lccc}
\toprule
\textbf{Property} & \textbf{Compound A} & \textbf{Compound B} & \textbf{Ideal Range} \\
\midrule
Molecular Weight (g/mol) & 347.84 & 389.86 & $<500$ \\
LogP & 3.77 & 3.49 & 1--5 \\
H-bond Donors & 1 & 0 & $\leq 5$ \\
H-bond Acceptors & 4 & 5 & $\leq 10$ \\
Rotatable Bonds & 5 & 4 & $\leq 10$ \\
TPSA (\AA$^2$) & 39.72 & 76.30 & 40--90 (optimal) \\
\bottomrule
\end{tabular}
\end{table}

\paragraph{Phase 2: Drug-likeness Rule Assessment}

Both compounds satisfy Lipinski's Rule of Five (0 violations) and Veber's rules for oral bioavailability. However, critical differences emerge in detailed analysis:

\begin{enumerate}
    \item \textbf{Polarity Balance:} Compound B's TPSA (76.30~\AA$^2$) falls within the optimal 40--90~\AA$^2$ range for oral absorption, while Compound A's TPSA (39.72~\AA$^2$) is slightly below optimal, potentially indicating solubility limitations.

    \item \textbf{3D Character:} Compound A contains two chiral centers (indicated by \texttt{@@} in SMILES), requiring stereoselective synthesis. Compound B has no stereocenters, simplifying manufacturing.

    \item \textbf{Structural Features:}
    \begin{itemize}
        \item Compound A: Morpholine-like ring, methoxy group, chloroaromatic
        \item Compound B: Sulfonamide group, isoxazole heterocycle, chloroaromatic
    \end{itemize}
\end{enumerate}

\paragraph{Phase 3: Reasoning and Decision}

\begin{quote}
``Candidate B has better polarity (TPSA in ideal range), likely better 3D character, no severe structural alerts, and no stereocenters simplifying synthesis. Candidate A's low TPSA might limit solubility and absorption, and chiral centers add complexity.''
\end{quote}

%% file: 9_appendix/matscibench_example.tex
\subsection{MatScibench Examples}
\label{sec:MatScibench Examples}

\subsubsection{Sample Task}

\paragraph{Problem Statement}

An electrochemical cell is constructed such that on one side a pure Zn electrode is in contact with a solution containing $\mathrm{Zn}^{2+}$ ions at a concentration of $10^{-2} \mathrm{M}$. The other cell half consists of a pure Pb electrode immersed in a solution of $\mathrm{Pb}^{2+}$ ions that has a concentration of $10^{-4} \mathrm{M}$. At what temperature will the potential between the two electrodes be +0.568 V ?

Reference:
"\text { Table 16.1 The Standard emf Series }"

\begin{tabular}{|l|l|}
\hline Electrode Reaction & Standard Electrode Potential, $V^0(\mathrm{~V})$ \\
\hline $\mathrm{Au}^{3+}+3 e^{-} \rightarrow \mathrm{Au}$ & +1.420 \\
\hline $\mathrm{O}_2+4 \mathrm{H}^{+}+4 e^{-} \rightarrow 2 \mathrm{H}_2 \mathrm{O}$ & +1.229 \\
\hline $\mathrm{Pt}^{2+}+2 e^{-} \rightarrow \mathrm{Pt}$ & $\sim+1.2$ \\
\hline $\mathrm{Ag}^{+}+e^{-} \rightarrow \mathrm{Ag}$ & +0.800 \\
\hline $\mathrm{Fe}^{3+}+e^{-} \rightarrow \mathrm{Fe}^{2+}$ & +0.771 \\
\hline $\mathrm{O}_2+2 \mathrm{H}_2 \mathrm{O}+4 e^{-} \rightarrow 4\left(\mathrm{OH}^{-}\right)$ & +0.401 \\
\hline $\mathrm{Cu}^{2+}+2 e^{-} \rightarrow \mathrm{Cu}$ & +0.340 \\
\hline $2 \mathrm{H}^{+}+2 e^{-} \rightarrow \mathrm{H}_2$ & 0.000 \\
\hline $\mathrm{Pb}^{2+}+2 e^{-} \rightarrow \mathrm{Pb}$ & -0.126 \\
\hline $\mathrm{Sn}^{2+}+2 e^{-} \rightarrow \mathrm{Sn}$ & -0.136 \\
\hline $\mathrm{Ni}^{2+}+2 e^{-} \rightarrow \mathrm{Ni}$ & -0.250 \\
\hline $\mathrm{Co}^{2+}+2 e^{-} \rightarrow \mathrm{Co}$ & -0.277 \\
\hline $\mathrm{Cd}^{2+}+2 e^{-} \rightarrow \mathrm{Cd}$ & -0.403 \\
\hline $\mathrm{Fe}^{2+}+2 e^{-} \rightarrow \mathrm{Fe}$ & -0.440 \\
\hline $\mathrm{Cr}^{3+}+3 e^{-} \rightarrow \mathrm{Cr}$ & -0.744 \\
\hline $\mathrm{Zn}^{2+}+2 e^{-} \rightarrow \mathrm{Zn}$ & -0.763 \\
\hline $\mathrm{Al}^{3+}+3 e^{-} \rightarrow \mathrm{Al}$ & -1.662 \\
\hline $\mathrm{Mg}^{2+}+2 e^{-} \rightarrow \mathrm{Mg}$ & -2.363 \\
\hline $\mathrm{Na}^{+}+e^{-} \rightarrow \mathrm{Na}$ & -2.714 \\
\hline $\mathrm{K}^{+}+e^{-} \rightarrow \mathrm{K}$ & -2.924 \\
\hline
\end{tabular}

\subsubsection{Sample Workflow}
\paragraph{S1-NexusAgent Report Agent} (In Brief):

The Nernst equation for the cell is:

\[
E = E^\circ - \frac{RT}{nF} \ln Q
\]

Given:
- \( E = 0.568 \, \text{V} \)
- \( E^\circ = 0.637 \, \text{V} \)
- \( R = 8.314 \, \text{J/(mol·K)} \)
- \( n = 2 \)
- \( F = 96485 \, \text{C/mol} \)
- \( Q = \frac{[\text{Zn}^{2+}]}{[\text{Pb}^{2+}]} = \frac{10^{-2}}{10^{-4}} = 100 \)

Substitute:

\[
0.568 = 0.637 - \frac{RT}{2F} \ln(100)
\]

\[
\frac{RT}{2F} \ln(100) = 0.637 - 0.568 = 0.069
\]

\[
T = \frac{0.069 \times 2F}{R \ln(100)}
\]

\[
\ln(100) = \ln(10^2) = 2 \ln(10) \approx 2 \times 2.302585 = 4.60517
\]

\[
T = \frac{0.069 \times 2 \times 96485}{8.314 \times 4.60517}
\]

\[
T \approx \frac{13314.93}{38.285} \approx 347.7 \, \text{K}
\]

\[
T \approx 347.7 - 273.15 \approx 74.6 \, ^\circ\text{C}
\]

Since the problem likely expects temperature in Kelvin (common in such calculations), the answer is:

\boxed{347.7}